\documentclass{article}

\PassOptionsToPackage{numbers, compress}{natbib}
 \usepackage[preprint]{neurips_2026}

\usepackage[utf8]{inputenc} 
\usepackage{algpseudocode}
\usepackage{algorithm}
\usepackage[T1]{fontenc}    
\usepackage{hyperref}       
\usepackage{url}            
\usepackage{booktabs}       
\usepackage{amsfonts}       
\usepackage{nicefrac}       
\usepackage{microtype}      
\usepackage{xcolor}         
\usepackage{bm}
\usepackage{amsmath}
\usepackage{graphicx}
\usepackage{wrapfig}
\usepackage{amssymb}
\usepackage{enumitem}
\usepackage{multirow}

\title{EMERGE: Resolution-Agnostic Point Cloud Generation with Equivariant Graph-Based Diffusion}

\author{%
  Ilias Mitsouras \ \ Nikolaos Chaidos \ \ Giorgos Stamou \ \ Athanasios Voulodimos\\
  National Technical University of Athens\\
  Athens, Greece\\
  \texttt{\{iliasmits,nchaidos\}@ails.ece.ntua.gr}\\
  \texttt{gstam@cs.ntua.gr}\\
  \texttt{thanosv@mail.ntua.gr}
}

\begin{document}

\maketitle

\begin{abstract}
 Point cloud generation has emerged as a crucial task for accurately capturing and reproducing the complexity of the physical world. However, existing generative approaches, predominantly relying on Transformers and Variational Autoencoders (VAEs), frequently ignore the continuous, non-grid topologies inherent to 3D spaces. Although the integration of graph-based structures has yielded significant benefits in related discriminative vision tasks, such geometric architectures remain noticeably absent from 3D generative modeling. To address this gap, we introduce \textit{\textbf{EMERGE}} (\textit{\textbf{E}quivariant \textbf{M}ulti-scal\textbf{e} GNN for \textbf{R}esolution-agnostic point cloud \textbf{GE}neration}), the first fully $SE(3)$-equivariant graph-based diffusion backbone explicitly designed to generate point clouds while preserving continuous spatial symmetries. Our framework bypasses the rigid resolution dependencies of standard generative pipelines, enabling zero-shot inference at multiple, arbitrary spatial resolutions.  Extensive empirical evaluations demonstrate that EMERGE achieves State-of-the-Art generation quality across standard metrics, while the strong inherent geometric inductive biases enable significantly faster training convergence compared to existing baseline methods.
\end{abstract}

\section{Introduction}

At its core, 3D point cloud generation is the process of learning the underlying geometric distribution of spatial data to sample and construct novel, high-fidelity 3D shapes. Recent advancements in the field have been predominantly driven by Diffusion Probabilistic Models (DPMs) \citep{LION, TIGER, PVD, DPM}, which successfully adapt sequential probabilistic denoising to point distributions. However, constrained by the computational overhead of the standard reverse diffusion process and the difficulty of modeling intricate structures, the field has quickly evolved to include techniques ranging from trajectory optimization and flow matching \citep{fastpc_straight_flows, not_so_optimal} to hybrid representations \citep{PVD} and latent space diffusion \citep{LION}.

Despite this rapid evolution, these continuous flow and diffusion-based models typically overlook the non-grid structures and continuous spatial symmetries that naturally exist in 3D environments. Formulating these symmetries as strict mathematical equivariances injects a powerful geometric inductive bias, which has been proven to significantly enhance representation learning in complex vision tasks \citep{strike_with_a_pose, spherical_images, equivariant_data_efficient_1}. Yet, even when current State-of-the-Art (SotA) models incorporate equivariance for predictive tasks, they are frequently restricted to discrete rotation groups \citep{epn,e2pn}, rendering them unsuitable for high-quality generative applications. Graph Neural Networks (GNNs) provide a natural framework to overcome this by dynamically aggregating local geometric features in unordered point clouds, driving SotA performance across discriminative 2D and 3D tasks \citep{visiongnn, dvhvisiongnn, equivariant_point_cloud}. Despite their success in these areas, graph-based architectures have remained noticeably absent from 3D generative modeling. Consequently, continuous 3D equivariance remains an underexplored approach in this domain.

To bridge this gap, we propose \textit{\textbf{EMERGE}} (\textit{\textbf{E}quivariant \textbf{M}ulti-Scal\textbf{e} GNN for \textbf{R}esolution-agnostic point cloud \textbf{GE}neration}), the first fully $SE(3)$-equivariant, graph-based diffusion model explicitly designed for point cloud generation (Figure~\ref{fig:architecture}). By extending the principles of Equivariant GNNs (EGNNs) into the 3D generative framework, our approach inherently respects continuous spatial symmetries while dynamically modeling intricate topologies. Furthermore, unlike existing generative architectures which remain fundamentally bottlenecked by the spatial resolution (i.e. number of points) observed during training, our framework introduces a powerful capability to natively generate point clouds at arbitrary inference resolutions. By introducing a parameter-free \textbf{\textit{distribution alignment}} technique, we dynamically negate density-induced distribution shifts, allowing our model to scale seamlessly at inference time, without any retraining, fine-tuning, or architectural modifications.\looseness=-1

Empirically, we demonstrate that EMERGE achieves SotA generation quality with substantially improved efficiency, requiring up to 10x fewer training epochs than existing methods. Our core contributions are as follows:
\begin{itemize}[topsep = 0pt, leftmargin = 0.7cm]
    \item We introduce \textbf{EMERGE}, the first fully $SE(3)$-equivariant diffusion backbone for 3D point cloud generation, natively integrating fundamental geometric inductive biases.
    \item We propose a novel \textbf{Continuous Canonical Frame Voxel Pooling} mechanism and a \textit{Global Invariant Feature Attention} module that efficiently capture multi-scale topologies while strictly preserving spatial symmetries.
    \item We prove that density-induced distribution shifts can be tackled with a novel zero-shot \textbf{distribution alignment} technique, enabling high-fidelity \textbf{3D super-resolution at arbitrary inference resolutions}.
    \item We evaluate our model both quantitatively and qualitatively against prior methods, achieving SotA performance while requiring significantly fewer training epochs.
\end{itemize}

\section{Related Work}

\paragraph{GNNs in 3D Vision} Graph Neural Networks have proven highly effective for modeling non-grid topologies in computer vision. Particularly in the 3D domain, where point clouds are fundamentally unordered and lack a regular grid, GNNs provide a natural framework to dynamically aggregate local geometric features. Consequently, they have been widely adopted for 3D downstream tasks such as segmentation \citep{DynamicGCN_segmentation, deepgcns_semantic_segmentation, 3dgcn_classification_segmentation}, object detection \citep{pointgnn_object_detection}, and classification \citep{Jiang2024DHGCNDH_class_seg, 3dgcn_classification_segmentation}. Despite their pervasive success in these discriminative tasks, graph-based architectures have remained noticeably absent from 3D generative modeling. In this work, we bridge this gap by introducing the first graph-based diffusion backbone explicitly designed to tackle the inherently more complex challenges of point cloud generation.\looseness=-1

\paragraph{Point Cloud Generation} Early methods utilizing continuous normalizing flows \citep{pointflow, softFlow} and VAEs~\citep{setVAE, rGAN} successfully modeled point distributions but faced scaling and quality bottlenecks. The field advanced significantly with Denoising Diffusion Probabilistic Models (DDPMs) \citep{denoising_diffusion_probabilistic_models}, which applied sequential denoising directly to point clouds \citep{DPM}. To mitigate high denoising latency and capture complex geometries, subsequent works quickly evolved to introduce trajectory optimization \citep{fastpc_straight_flows, not_so_optimal} and hybrid point-voxel architectures like PVD \citep{PVD}. Building on the need for scalable representations, Latent Point Diffusion Models (LION) \citep{LION} map point clouds into a hierarchical latent space using a VAE, performing the diffusion process entirely within this highly expressive space. More recently, TIGER \citep{TIGER} introduced a time-varying denoising model that adaptively allocates its local and global representational power depending on the stage of the generation process. While these methods have achieved impressive generative capabilities, they largely ignore explicit non-grid topologies and continuous spatial symmetries inherent in the 3D space. Our proposed method addresses this fundamental limitation by introducing a graph-based backbone that enforces strict $SE(3)$-equivariance, while enabling zero-shot super-resolution generation at arbitrary resolutions during inference.\looseness=-1

\paragraph{2D/3D Equivariant Models} Embedding equivariance or invariance directly into a backbone model endows it with strong inductive biases regarding specific input transformations. This property has been proven to significantly enhance representation learning, outperforming standard architectures and heavy data augmentation in complex vision tasks \citep{strike_with_a_pose, spherical_images}. Beyond standard computer vision, 3D-informed architectures like Equivariant GNNs \citep{egnn} have been fundamental in modeling physical dynamics \citep{egnn_normalizing_flows, egnn_local_frames}, as well as in both predictive and generative tasks for chemical compounds \citep{egnn, egnn_topological, egnn_edm, digress}. Although certain models have incorporated limited equivariance (typically restricted to a discrete group of rotations) for downstream point cloud analysis \citep{epn, e2pn, equivariant_point_cloud}, 3D equivariance has remained entirely absent from current generative models; the need for a universal, continuous geometric inductive bias in point cloud synthesis directly motivates our proposed architecture.

\section{Preliminaries}
\label{sec:preliminaries}
\paragraph{Diffusion Models} Diffusion models are a class of generative probabilistic models that operate by iteratively corrupting the original data with noise during the forward diffusion process and subsequently learning to reverse this process to synthesize novel samples \cite{diffusion_non_equillibrium, denoising_diffusion_probabilistic_models}. Given a point cloud $\mathbf{X}_{0}  \in \mathbb{R}^{N \times 3}$ sampled from an unknown underlying data distribution $q(\mathbf{X}_{0})$, where $N$ is the total number of points, the forward process is modeled as a predefined Markov chain that progressively adds Gaussian noise over $t=1,...,T$ discrete timesteps, so that  $\mathbf{X}_{T}\sim \mathcal{N}(\mathbf{0}, \mathbf{I})$.  Denoising Diffusion Probabilistic Models are trained to approximate the intractable true posterior distribution $q(\mathbf{X}_{t-1}| \mathbf{X}_{t})$. Following~\cite{denoising_diffusion_probabilistic_models}, their simplified training objective is formulated as:
 \begin{equation}
     \mathcal{L}_{simple}  =  \mathbb{E}_{t\sim \mathcal{U}[1,T], \mathbf{X}_{0}\sim q(\mathbf{X}_{0}), \boldsymbol{\epsilon}\sim \mathcal{N}(\mathbf{0}, \mathbf{I})}\left[\| \boldsymbol{\epsilon} -  \boldsymbol{\epsilon}_\theta(\sqrt{\bar{\alpha}_t} \mathbf{X}_{0}  + \sqrt{1- \bar{\alpha}_t} \boldsymbol{\epsilon}, t)\|_{2}^{2}\right], \label{diffusion_noise_loss}
\end{equation}
where $\bar{\alpha}_t = \prod_{s = 1}^{t}(1 - \beta_s)$, $\beta_t$ is the variance schedule and $\boldsymbol{\epsilon}_{\theta}$  is a neural network trained to predict the added noise $\boldsymbol{\epsilon}$.

\paragraph{$\boldsymbol{SE(3)}$-Equivariance and Invariance}
Formally, the Special Euclidean group $SE(3)$ describes transformations consisting of a rotation matrix $\mathbf{R}\in SO(3)$ and a translation vector $\mathbf{t}\in\mathbb{R}^{3\times 1}$. A function $f_{\theta}:\mathbb{R}^{N\times3}\rightarrow\mathbb{R}^{N\times3}$ processing a point cloud $\mathbf{X}_0$ is considered strictly $SE(3)$-equivariant if applying the transformation to the input results in an identically transformed output:
\begin{equation}
    f_{\theta}(\mathbf{X}_{0} \mathbf{R}+\mathbf{1}_{N}\mathbf{t}^{T})=f_{\theta}(\mathbf{X}_{0})\mathbf{R}+\mathbf{1}_{N}\mathbf{t}^{T},
\end{equation}
where $\mathbf{1}_{N}\in\mathbb{R}^{N\times 1}$ is a vector of ones. In contrast, a function $h_{\phi}:\mathbb{R}^{N\times3}\rightarrow\mathbb{R}^{N\times 3}$ is considered strictly $SE(3)$-invariant if its output remains unaffected by spatial transformations of the input, $h_{\phi}(\mathbf{X}_{0}\mathbf{R}+\mathbf{1}_{N}\mathbf{t}^{T})=h_{\phi}(\mathbf{X}_{0})$. For brevity, all subsequent references to equivariance or invariance in this work denote $SE(3)$-equivariance or invariance, respectively, unless otherwise specified, and we omit the explicit $\mathbf{1}_{N}$ vector when applying translations by assuming standard dimension broadcasting.

\section{Methodology}
\label{section_methodology}

\subsection{EMERGE Architecture}
Given a noisy point cloud $\mathbf{X}_t \in \mathbb{R}^{N\times 3}$ at timestep $t$ of the forward diffusion process, our framework processes the geometry through a sequence of $N_{B}$ identical blocks, as shown in Figure \ref{fig:architecture}. To balance local geometric modeling with global structural coherence, each block alternates between two distinct processing phases:
\begin{enumerate}[topsep = 0pt, leftmargin = 0.7cm]
\item \textbf{Local Processing:} A hierarchical refinement phase that aggregates local features at varying spatial resolutions using a Multi-Scale EGNN and a novel Canonical Frame Voxel Pooling mechanism.
\item \textbf{Global Processing:} A macro-level phase that injects global shape context into the node features via an efficient Global Invariant Feature Attention module.
\end{enumerate}
Repeating this sequential processing progressively refines the geometric representation, ultimately yielding the denoised spatial coordinates $f_\theta(\mathbf{X}_t, t)$ while strictly preserving the $SE(3)$ symmetries of the 3D domain. The following subsections detail the local and global mechanisms within each block.\looseness=-1

\subsection{Local Processing: Multi-Scale EGNN}
\label{sec:multiscale_egnn}
To effectively model the complex geometry of 3D point clouds within a diffusion framework, our architecture must inherently respect the continuous spatial symmetries of the 3D domain. While prior works on 3D predictive tasks such as E2PN \cite{e2pn}, have incorporated equivariance, they are often restricted to discrete subgroups of rotations (e.g., 16 predefined viewing angles in \cite{e2pn}, 60 in~\cite{epn}), which inherently limits their representational capacity and can introduce discretization artifacts, disqualifying these approaches for use in generative tasks. To overcome this, our backbone extends EGNNs \cite{egnn} to achieve continuous, universal $SE(3)$-equivariance without requiring expensive data augmentation or restricting the model to specific spatial orientations. \looseness=-1

\begin{figure}[t]
    \centering
    \includegraphics[width=\textwidth]{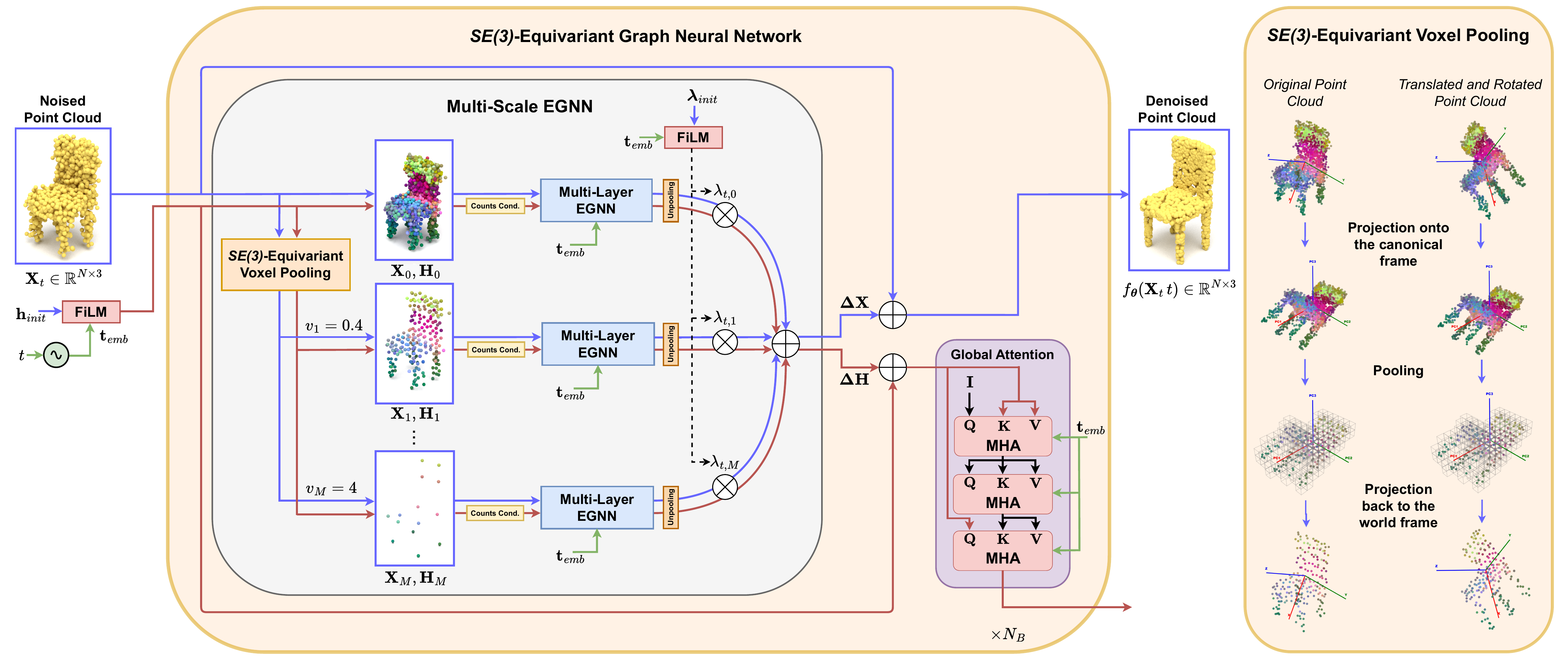}
    \caption{\textbf{Overview of the EMERGE architecture}. Given a noisy point cloud $\mathbf{X}_t$ at timestep $t$, our framework processes the geometry through $N_{\textit{B}}$ sequential blocks. Within each block, a Multi-Scale EGNN aggregates local features at varying spatial resolutions using our novel $SE(3)$-Equivariant Voxel Pooling (detailed right), processing each of the $M+1$ scales through a Multi-Layer EGNN consisting of $N_E$ sequential message-passing layers. The hierarchical local updates are passed to the Global Invariant Feature Attention module, which efficiently captures macro-level shape context via inducing points. The model outputs the prediction of the denoised point cloud $f_\theta(\mathbf{X}_t, t)$ while strictly preserving continuous spatial symmetries.}
    \label{fig:architecture}
\end{figure}

\paragraph{Feature Initialization} We begin by initializing a globally shared, learnable feature vector $\mathbf{h}_{init} \in~\mathbb{R}^D$. To condition the network on the continuous diffusion timestep $t$, we generate a standard sinusoidal embedding $\mathbf{t}_{emb} \in \mathbb{R}^D$:
\begin{equation}
    \mathbf{t}_{emb}^{(2n)} = \sin\left(\frac{t}{10,000^{2n/D}}\right), \quad \mathbf{t}_{emb}^{(2n+1)} = \cos\left(\frac{t}{10,000^{2n/D}}\right),
\end{equation}
where $n$ denotes the dimension index. To construct the initial per-node feature embedding $\mathbf{h}_{t,i}^0 \in \mathbb{R}^D$ for each point $i \in \{1,...,N\}$, this shared vector is copied across all points and conditioned on the continuous diffusion timestep $t$ through a Feature-wise Linear Modulation (FiLM) layer \citep{film} (further details are provided in Appendix \ref{appendix_film_layer}):\looseness=-1
\begin{equation}
\mathbf{h}_{t,i}^0 = \text{FiLM}(\mathbf{h}_{init}, \mathbf{t}_{emb}).    
\end{equation}

\begin{wrapfigure}{r}{0.35\textwidth}
    \vspace{-25pt}
    \centering
    \includegraphics[width=\linewidth]{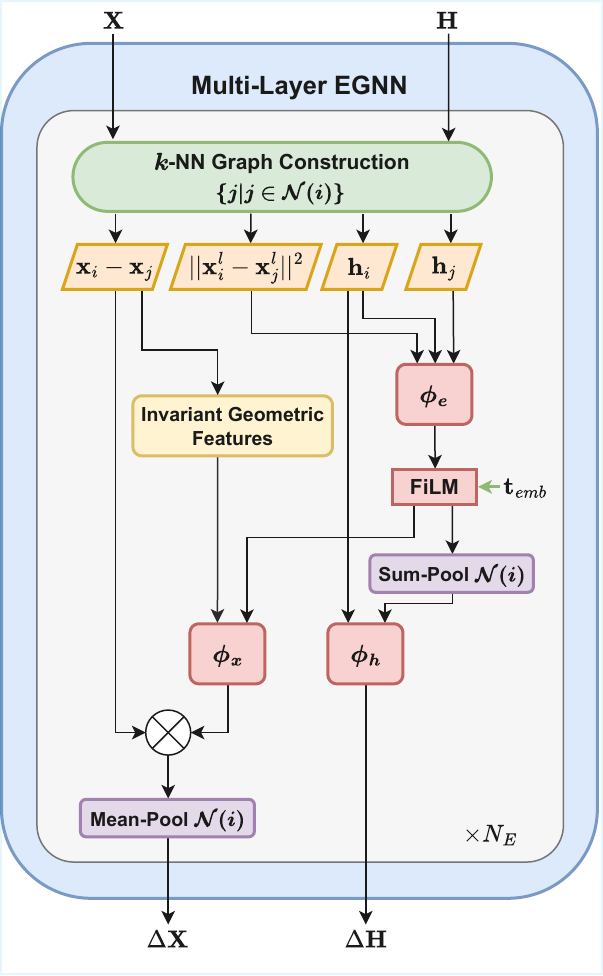}
    \vspace*{-20pt}
    \caption{\textbf{Multi-Layer EGNN Block Architecture}.}
    \label{fig:multilayer_egnn_architecture}
    \vspace{-20pt}
\end{wrapfigure}

\paragraph{Multi-Layer EGNN} The standard EGNN \citep{egnn} iteratively updates equivariant coordinates and invariant features. To effectively scale this for the deep generative process, we recast the updates as residual functions ($\Delta \mathbf{x}^l, \Delta \mathbf{h}^l$) across multiple layers. Because point clouds lack a regular grid, we dynamically construct a local neighborhood graph at each layer $l$ by computing the $k$-Nearest Neighbors ($k$-NN) for every point, based on their current spatial coordinates $\mathbf{x}_{t,i}^{l}$, using a fixed parameter $k$ across the entire network. For notational clarity, we drop the timestep $t$ from all intermediate activations and coordinates. Furthermore, we inject the diffusion timestep directly into the edge messages via intermediate FiLM layers. For a given layer $l\in\{1, \ldots , N_{E}\}$, the updates are formulated as (Figure \ref{fig:multilayer_egnn_architecture}):\looseness=-1
\begin{equation}
    \mathbf{m}_{ij} = \phi_e\left(\mathbf{h}_i^l, \mathbf{h}_j^l, ||\mathbf{x}_i^l - \mathbf{x}_j^l||^2\right),\quad
\tilde{\mathbf{m}}_{ij} = \text{FiLM}(\mathbf{m}_{ij}, \mathbf{t}_{emb}),
\end{equation}
\begin{equation}
    \Delta\mathbf{x}_i^l = \frac{1}{|\mathcal{N}(i)|} \sum_{j \in \mathcal{N}(i)} (\mathbf{x}_i^l - \mathbf{x}_j^l) \phi_x\left(\tilde{\mathbf{m}}_{ij}, \mathbf{f}_{inv, i}\right),\quad \Delta\mathbf{h}_i^l = \phi_h\left(\mathbf{h}_i^l, \sum_{j \in \mathcal{N}(i)} \tilde{\mathbf{m}}_{ij}\right).\label{eq:egnn_f_inv}
\end{equation}
The features and coordinates are then advanced via residual addition: $\mathbf{x}_i^{l+1} = \mathbf{x}_i^l + \Delta\mathbf{x}_i^l$ and $\mathbf{h}_i^{l+1} = \mathbf{h}_i^l + \Delta\mathbf{h}_i^l$. Here, we model the $\phi_e$, $\phi_x$, and $\phi_h$ functions as MLPs, and $\mathbf{f}_{inv, i}$ represents our novel injection of local invariant geometric features.

\paragraph{Invariant Geometric Features} A key limitation of processing raw relative distances is the loss of descriptive local geometric context. While pairwise distances are informative, they do not explicitly capture higher-level geometric properties, such as the local \textit{curvature}, \textit{linearity}, or \textit{planarity} of the point cloud at a specific neighborhood. To enrich the spatial messaging $\phi_x$, we compute a 13-dimensional feature vector $\mathbf{f}_{inv, i}$ describing these spatial characteristics around the neighborhood of point $i$ at layer $l$, alongside absolute spatial moments. These features are derived from the eigenvalues and eigenvectors of the neighborhood's covariance matrix. Intuitively, these geometric properties are inherently $SE(3)$-invariant; the curvature, linearity, and planarity of a local patch remain identical regardless of how the overall point cloud is rotated or translated. 

The foundational EGNN architecture is already proven to be $SE(3)$-equivariant \citep{egnn}. Injecting these rich geometric features does not violate the network's strict equivariance. Informally, any SE(3) transformation applied to a local neighborhood simply rotates its covariance matrix, leaving its intrinsic eigenvalues perfectly unchanged. While the eigenvectors rotate alongside the input points, projecting the relative distance vectors onto these rotated eigenvectors naturally cancels out the rotation, yielding purely invariant scalars. Because $\mathbf{f}_{inv, i}$ consists exclusively of these invariant scalars, the coordinate update $\Delta\mathbf{x}_i^l$ remains a linear combination of equivariant vectors weighted by invariant scalars, completely preserving the $SE(3)$-equivariance of the network. A formal mathematical proof, along with a detailed breakdown of the 13 invariant features, is provided in Appendix \ref{invariant-features-appendix}.\looseness=-1

\paragraph{$\boldsymbol{SE(3)}$-Equivariant Voxel Pooling} To generate high-fidelity 3D shapes, the network must perceive geometry at multiple hierarchical levels \citep{PVD, LION}, capturing both fine surface details and broad structural topologies. However, standard voxel-pooling maps points into a fixed, axis-aligned 3D grid; if the point cloud rotates, the points fall into different cells, inherently breaking equivariance. To maintain strict spatial symmetries without restricting our diffusion backbone to predefined discrete viewing angles, we introduce \textbf{Canonical Frame Voxel Pooling}, a technique that enables universal, continuously $SE(3)$-equivariant point aggregation.\looseness=-1

Given a point cloud $\mathbf{X} \in \mathbb{R}^{N \times 3}$, we first mean-center the coordinates using their centroid $\boldsymbol{\mu}$ to achieve translation invariance: $\mathbf{X}_c = \mathbf{X} - \boldsymbol{\mu}$. Next, we compute the covariance matrix $\mathbf{C} = \mathbf{X}_c^T \mathbf{X}_c$ and extract a unique, right-handed orthogonal canonical frame $\mathbf{V}^* \in SO(3)$ via its eigenvectors, resolving inherent sign and chirality ambiguities (Appendix \ref{equivariant-voxel-pooling-appendix}). Intuitively, any rotation of the input simply rotates $\mathbf{V}^*$ by the exact same amount (see Figure \ref{fig:architecture}). Consequently, projecting the points onto this frame ($\mathbf{Y} = \mathbf{X}_c \mathbf{V}^*$) perfectly cancels the transformation, yielding strictly invariant coordinates aligned to the object's principal axes. Within this invariant space, we safely apply standard spatial voxel pooling with grid size $m$ to obtain pooled nodes $\mathbf{Y}_{m} = \text{VoxelPool}(\mathbf{Y}, m)$ alongside mean-pooled node features $\mathbf{H}_{m}$. Finally, the pooled coordinates are unprojected back into the original world space, ensuring they rotate in perfect sync with the input: \looseness=-1
\begin{equation}
\mathbf{X}_{m} = \mathbf{Y}_{m} (\mathbf{V}^*)^T + \boldsymbol{\mu}.
\end{equation}
\vspace*{-2\baselineskip}

\paragraph{Counts-Conditioning} When mapping continuous points to discrete voxels, varying node densities emerge (some voxels aggregate many points, while others capture only a few), which is a critical characteristic of the point cloud's underlying local structure. To ensure the network is aware of this structural density at different scales, we compute the point count $c_n$ for each voxel-pooled node $n$ and use it to condition the node features $\mathbf{H}_{m}$ via a FiLM layer (Appendix \ref{appendix_film_layer}). Because the voxel grid is constructed inside the canonical space $\mathbf{Y}$, the spatial boundaries of the grid are rotation and translation invariant. Therefore, the number of points falling into each voxel remains unchanged under any SE(3) transformation, meaning $c_n$ is an invariant scalar.

\paragraph{Unpooling and Scale Aggregation} Let  $\mathbf{V}_{vox} = \{v_0, v_1, \ldots, v_M\} \subset \mathbb{R}_{>0}^{M+1}$ denote the $M$ distinct voxel sizes alongside the unpooled base resolution ($v_0$), yielding coordinate updates $\Delta \mathbf{X}_m$ and feature updates $\Delta \mathbf{H}_m$ for each scale $v_m$. During unpooling, we simply copy the voxel output to each of its assigned constituent points, so that $\Delta \mathbf{X}_m \in \mathbb{R}^{N \times 3}$ and $\Delta \mathbf{H}_m \in \mathbb{R}^{N \times D}$. Different scales hold varying importance depending on the current stage of the model; fine details are critical during early diffusion timesteps for noise removal, while global structures are dominant at later timesteps for macro-shape formation. Inspired by similar scale-mixing approaches \citep{immpnn}, we introduce adaptable coefficients $\boldsymbol{\lambda}_{t} = \text{FiLM}(\boldsymbol{\lambda}_{init}, \mathbf{t}_{emb}) \in \mathbb{R}^{M+1}$ to aggregate the per-scale predictions, with a learnable base vector $\boldsymbol{\lambda}_{init}$ conditioned dynamically on the diffusion timestep:
\begin{equation}
    \Delta \mathbf{X} = \sum_{m=0}^M \lambda_{t,m} \Delta \mathbf{X}_m, \qquad  \Delta \mathbf{H} = \sum_{m=0}^M \lambda_{t,m} \Delta \mathbf{H}_m.
\end{equation}
Since the coefficients $\mathbf{\lambda}_{t, m}$ are invariant scalars, then the linear combination with the equivariant coordinate updates $\Delta \mathbf{X}_m$ and the invariant features $\Delta \mathbf{H}_m$ preserve their respective $SE(3)$ symmetries.

\subsection{Global Processing: Invariant Feature Attention}
To efficiently capture global shape context and complement the local inductive biases of our graph-based messaging, we incorporate a global attention mechanism. We build upon the highly scalable \textit{Inducing Point Attention} paradigm \citep{set_transformer} and extend it for our denoising diffusion backbone. This approach is advantageous for its linear computational complexity relative to $N$ and because it inherently preserves the $SE(3)$-invariant properties of the hidden node features.

Standard Multi-Head Attention (MHA) across all $N$ points is computationally prohibitive for dense point clouds. Instead, we introduce a small set of learnable inducing points $\mathbf{I} \in \mathbb{R}^{C \times D}$, where $C \ll N$. The global attention is computed in three sequential stages:
\begin{enumerate}[topsep = 0pt, leftmargin = 0.7cm]
    \item \textit{Compression:} A cross-attention layer where the inducing points $\mathbf{I}$ act as queries to aggregate information from all node features $\mathbf{H}$ (which act as keys and values). 
    \item \textit{Processing:} A self-attention layer applied exclusively over the inducing points $\mathbf{I}$ to capture the global structural context.
    \item \textit{Broadcasting:} A final cross-attention layer where the original node features $\mathbf{H}$ query the updated inducing points to retrieve the globally-aware context.
\end{enumerate}

Because the entire module consists exclusively of, either point-wise transformations (e.g., FFNs, non-linearities, and layer normalizations), or permutation-equivariant aggregations (MHA layer without positional encoding) applied directly to the invariant node features $\mathbf{H}$, the resulting updated features retain their spatial invariance without requiring additional modifications to the attention mechanism. \looseness=-1

To make this module suitable for the reverse diffusion process, we modulate the feature representations at each stage based on the current timestep $t$. We replace standard Layer Normalization with Adaptive Layer Normalization (adaLN) \citep{adaln}, a technique widely used in diffusion models. The scale and shift parameters for the adaLN layers are regressed directly from the timestep embedding $\mathbf{t}_{emb}$.

\section{Experiments}\label{sec:experiments}

\textbf{Datasets} \; To be consistent with previous work, we utilize ShapeNetv2 \cite{shapenet} as the primary dataset for training and evaluating our model, specifically using the preprocessing pipeline introduced by PointFlow \cite{pointflow}. In particular, we train our model on the three widely adopted categories: \textit{airplane}, \textit{chair}, and \textit{car}, containing 2,832, 4,612, and 2,458 shapes, respectively. For training, we apply per-shape normalization while each shape is sampled to contain exactly $N = 2,048$ points.

\textbf{Implementation Details} \; Our architecture utilizes $N_{\textit{B}} = 6$ main blocks, each interleaving a local Multi-Scale EGNN with a Global Inducing Point Attention module. Within the Multi-Scale EGNN, each of the $7$ resolution branches (the base resolution plus $M=6$ voxel scales) is independently processed by a Multi-Layer EGNN consisting of $N_E = 2$ sequential message-passing layers ($k=5$). Full hyperparameters, including the exact multi-resolution voxel pooling array, are detailed in Appendix \ref{app:implementation_details}.\looseness=-1

\textbf{Training} \; The model is trained using a continuous diffusion process over $T=1000$ timesteps, with a linear noise schedule ranging from $\beta_{start}=10^{-4}$ to $\beta_{end}=0.02$. Following standard practices, our model predicts the clean data $\mathbf{X}_0$ using the equivalent reparameterized form of the simplified diffusion objective in Eq. (\ref{diffusion_noise_loss}). To mitigate training instabilities and alleviate slow convergence across timesteps, we apply the Min-SNR weighting strategy \cite{Hang2023EfficientDT} with $\gamma=5$ (full formulation provided in Appendix \ref{app:min_snr}).\looseness=-1

\textbf{Evaluation Metrics} \; Following previous works \cite{pointflow, LION, TIGER}, to quantitatively assess the performance of our proposed model, we employ 1-Nearest Neighbor Accuracy (1-NNA) \cite{pointflow} as our primary evaluation metric, calculated using both Chamfer Distance (CD) and Earth Mover's Distance (EMD). This metric computes the accuracy of a leave-one-out 1-NN classifier tasked with distinguishing between generated and ground-truth shapes. An ideal accuracy of $50\%$ indicates that the generated distribution is entirely indistinguishable from the real one, reflecting optimal generation quality. Since EMERGE is $SE(3)$-equivariant, we also isolate structural quality by automatically aligning the principal axes of all shapes via PCA (more details in Appendix \ref{app:implementation_details}).

\begin{table}[t]
\def\arraystretch{1.15}
\centering
\small
\caption{Evaluation metrics (1-NNA $\downarrow$) utilizing both Chamfer Distance (CD) and Earth Mover's Distance (EMD). Lower values indicate better generation quality, with 50\% representing the theoretical optimum. We evaluate across the standard dataset splitting strategy (upper section) and the LION dataset splitting strategy \cite{LION} (lower section). Baseline results are taken from \citep{LION}.}
\begin{tabular}{lcccccc}
\toprule
 & \multicolumn{2}{c}{\textbf{Airplane}} & \multicolumn{2}{c}{\textbf{Chair}} & \multicolumn{2}{c}{\textbf{Car}} \\
\cmidrule(lr){2-3} \cmidrule(lr){4-5} \cmidrule(lr){6-7}
\textbf{Model} & CD ($\downarrow$)& EMD ($\downarrow$)& CD ($\downarrow$)& EMD ($\downarrow$)& CD ($\downarrow$)& EMD ($\downarrow$)\\
\midrule
\midrule
r-GAN \cite{rGAN} & 98.40 & 96.79 & 83.69 & 99.70 & 94.46 & 99.01 \\
l-GAN (CD) \cite{rGAN} & 87.30 & 93.95 & 68.58 & 83.84 & 66.49 & 88.78 \\
l-GAN (EMD) \cite{rGAN} &  89.49 & 76.91 & 71.90 & 64.65 & 71.16 & 66.19 \\
PointFlow \cite{pointflow}  & 75.68 & 70.74 & 62.84 & 60.57 & 58.10 & 56.52 \\
DPF-Net \cite{klokov20eccv} & 75.18 & 65.55 & 62.00 & 58.53 & 62.35 & 54.48 \\
ShapeGF \cite{ShapeGF} & 80.00 & 76.17 & 68.96 & 65.48 & 63.20 & 56.53\\
SoftFlow \cite{softFlow} & 76.05 & 65.80 & 59.21 & 60.05 & 64.77 & 60.09 \\
SetVAE \cite{setVAE} & 76.54 & 67.65 & 58.84 & 60.57 & 59.95 & 59.94 \\
DPM \cite{DPM}  & 76.42 & 86.91 & 60.05 & 74.77 & 68.89 & 79.97 \\
PVD  \cite{PVD} & 73.82 & 64.81 & 56.26 & 53.32 & 54.55 & 53.83 \\
TIGER \cite{TIGER} & 71.85 & 55.82 & 54.61 & 52.71 & 54.31 & 52.24 \\
\textbf{EMERGE} & \textbf{55.93} & \textbf{53.33} & \textbf{54.31} & \textbf{52.42} & \textbf{53.55}& \textbf{50.71}\\
\midrule
LION \cite{LION} & 67.41 & 61.23 & 53.70 & 52.34 & 53.41 & 51.14\\
TIGER \cite{TIGER} & 67.21 & 56.26 & 54.32 & 51.71 & 54.12 & \textbf{50.24}\\
\textbf{EMERGE} & \textbf{56.17}& \textbf{54.69}& \textbf{53.55}&  \textbf{51.44}& \textbf{52.41}& 50.71\\
\bottomrule
\bottomrule 
\end{tabular}
\label{quantitative_results_table}
\end{table}

\begin{figure}[!b]
    \centering
    \includegraphics[width=\textwidth]{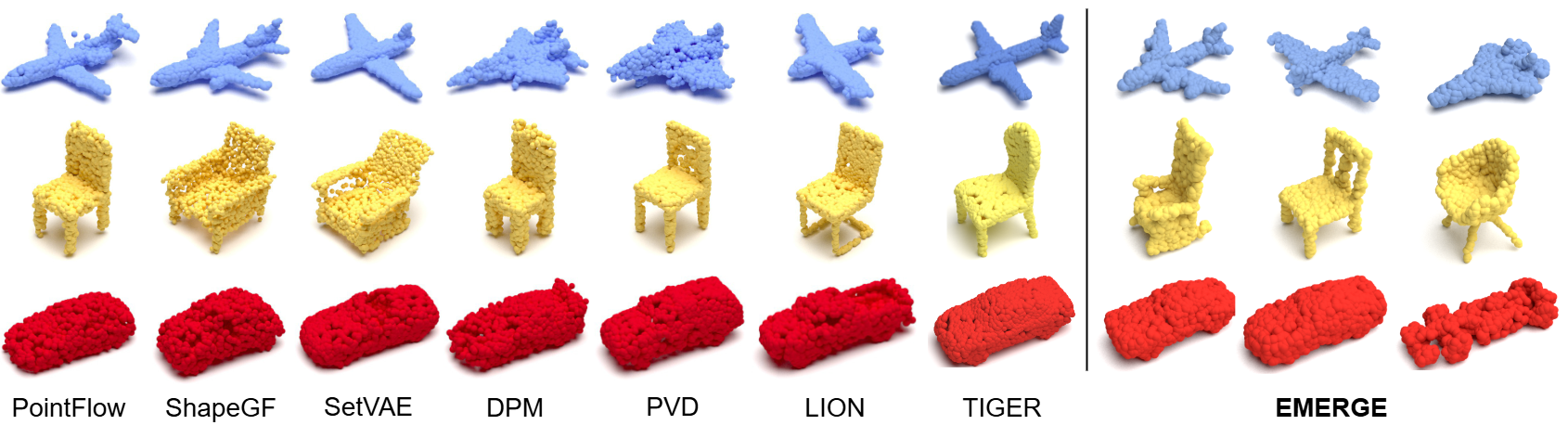}
    \caption{Samples generated by EMERGE alongside various baselines for the \textit{airplane}, \textit{chair}, and \textit{car} categories. Each point cloud consists of $N = 2,048$ points.}
    \label{fig:main_qual}
\end{figure}

\textbf{Results} \; All samples for quantitative and qualitative evaluation are generated using 100 Denoising Diffusion Implicit Models (DDIM) \citep{ddim} steps. Quantitative results, summarized in Table \ref{quantitative_results_table}, demonstrate that our model achieves SotA performance. Under the standard splitting strategy, we exceed all baseline metrics, and under the LION splitting strategy, we lead in 5 out of 6 metrics. Notably, our model approaches the ideal 50\% 1-NNA mark with 50.71\% EMD for cars, 53.33\% EMD for airplanes, and 52.42\% EMD for chairs. Furthermore, injecting equivariance into the backbone historically leads to dramatic data and training efficiency \cite{equivariant_data_efficient_1, equivariant_data_efficient_2, equivariant_data_efficient_3}. This is strongly evident in our framework: while recent diffusion baselines require extensive training regimes (such as LION needing 32,000 total epochs, 24,000 for the latent DPM plus 8,000 for the VAE), our model converges much earlier, requiring only 3,100 epochs for airplanes, 2,600 for chairs, and 7,000 for cars. Qualitatively, as shown in Figure \ref{fig:main_qual}, EMERGE consistently produces high-fidelity shapes with clear local surfaces and global structural integrity compared to other methods.

\begin{figure}[t]
    \centering
    \includegraphics[width=\textwidth]{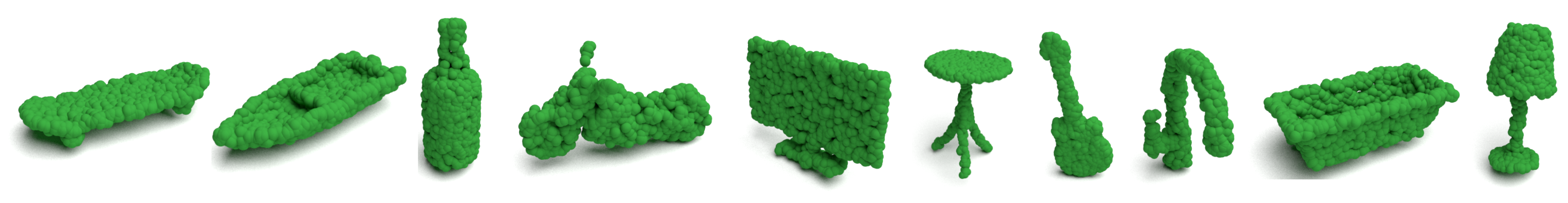}
    \caption{Samples generated by EMERGE for all 55 categories of ShapeNet ($N = 2,048$).}
    \label{fig:55_class}
    \vspace{-\baselineskip}
\end{figure}

Finally, to demonstrate the generalizability of our framework, we trained EMERGE on all 55 different categories from ShapeNet. As visualized in Figure \ref{fig:55_class}, our model successfully generates highly diverse and structurally complex point clouds without relying on explicit class conditioning.\looseness=-1

\section{Zero-Shot Super-Resolution}\label{sec:zero_shot_multiresolution}

\begin{figure}[b]
    \vspace{-\baselineskip}
    \centering
    \includegraphics[width = \textwidth]{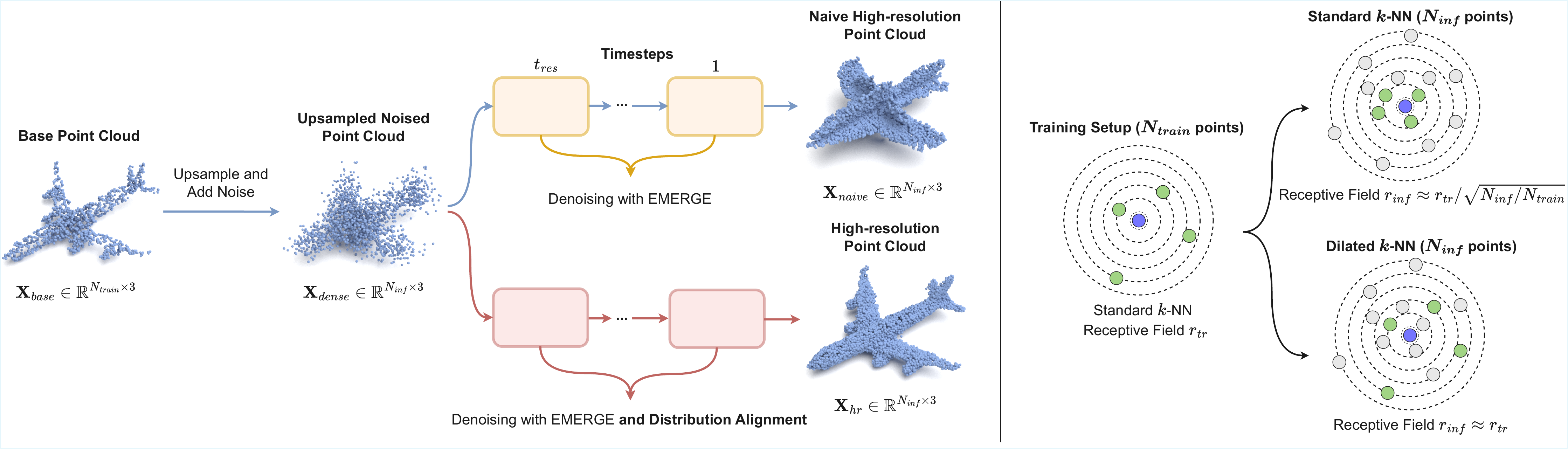}
    \caption{\textbf{Zero-shot super-resolution with EMERGE}. \textbf{(Left)} Naive denoising at high resolutions introduces severe distribution shifts, causing structural failure (top). Our parameter-free distribution alignment technique negates these shifts, enabling high-fidelity super-resolution without retraining (bottom). \textbf{(Right)} We resolve density-induced receptive field shrinkage by introducing a Dilated $k$-NN graph, perfectly preserving the receptive field $r_{train}$ at arbitrary inference resolutions.}
    \label{fig:multires}
\end{figure}

Standard diffusion-based architectures face severe distribution shifts when evaluated at higher point densities, necessitating full retraining for varying inference resolutions. Although hierarchical VAEs like SetVAE \citep{setVAE} have attempted resolution-aware modeling, their VAE backbones fundamentally bottlenecked their generation quality and geometric expressivity. To the best of our knowledge, EMERGE is the first high-quality diffusion-based framework to introduce an unprecedented, zero-shot capability to natively super-resolve point clouds at arbitrary inference densities ($N_{inf} > N_{train}$). We show that density-induced distribution shifts are attributed to exactly two sub-components and introduce a parameter-free alignment technique to dynamically negate them during inference (Figure~\ref{fig:multires}). This exact alignment is only possible because our model is fundamentally a distance-based GNN, unlike prior frameworks \citep{LION, TIGER}.\looseness=-1

The two primary bottlenecks that occur when performing inference at $N_{inf} > N_{train}$ and their respective alignment propositions are detailed below.

\textbf{Voxel Count Distribution Shift \& Alignment} \ At higher resolutions, the expected points per voxel increases proportionally with the density ratio $\rho = N_{inf} / N_{train}$, pushing the inputs of the count-conditioning FiLM layers entirely out of the training distribution.

\textbf{Proposition 1 (Point-Count Alignment)} \ \textit{To preserve the expected input distribution of the count-conditioning FiLM layers, the raw inference voxel counts $c_n(N_{inf})$ must be linearly scaled by the inverse density ratio prior to normalization: $\hat{c}_n = c_n(N_{inf}) \cdot (N_{train} / N_{inf})$. (Proof in Appendix \ref{appendix_scale_count_alignment}).}\looseness=-1


\textbf{Receptive Field Shrinkage \& Alignment} \ Assuming points are approximately uniformly distributed over a locally smooth 2D surface manifold, the spatial radius $r_{k}$ of the EGNN's ($k$-NN) graph scales as $r_{k} \propto \sqrt{k / N}$. Consequently, querying $k$ neighbors at $N_{inf}$ artificially shrinks the network's spatial radius by a factor of $\sqrt{\rho}$, inducing a mismatch with the receptive field observed during training.\looseness=-1

\textbf{Proposition 2 (Dilated $\boldsymbol{k}$-NN Receptive Field Matching)} \ \textit{Let $d = \lfloor N_{inf} / N_{train} \rfloor$ be the spatial dilation factor. A dilated k-NN graph querying $k_{ext} = k \cdot d$ neighbors and subsampling every $d$-th neighbor perfectly preserves both the spatial receptive field radius $r_{train}$ and the message-passing cardinality $k$. (Proof in Appendix \ref{appendix_dilated_knn}).}


With these two bottlenecks resolved after applying the parameter-free modifications of Propositions~1 and 2, the rest of our architecture remains inherently immune to density shifts. Because the EGNN coordinate updates rely strictly on localized $k$-cardinal aggregations and the global attention mechanism scales natively via a fixed number of inducing points, the overall feature magnitudes remain entirely invariant to the raw point count (detailed analysis in Appendix \ref{appendix_architecture_immunity}).\looseness=-1

\begin{figure}[t]
    \centering
    \includegraphics[width = \textwidth]{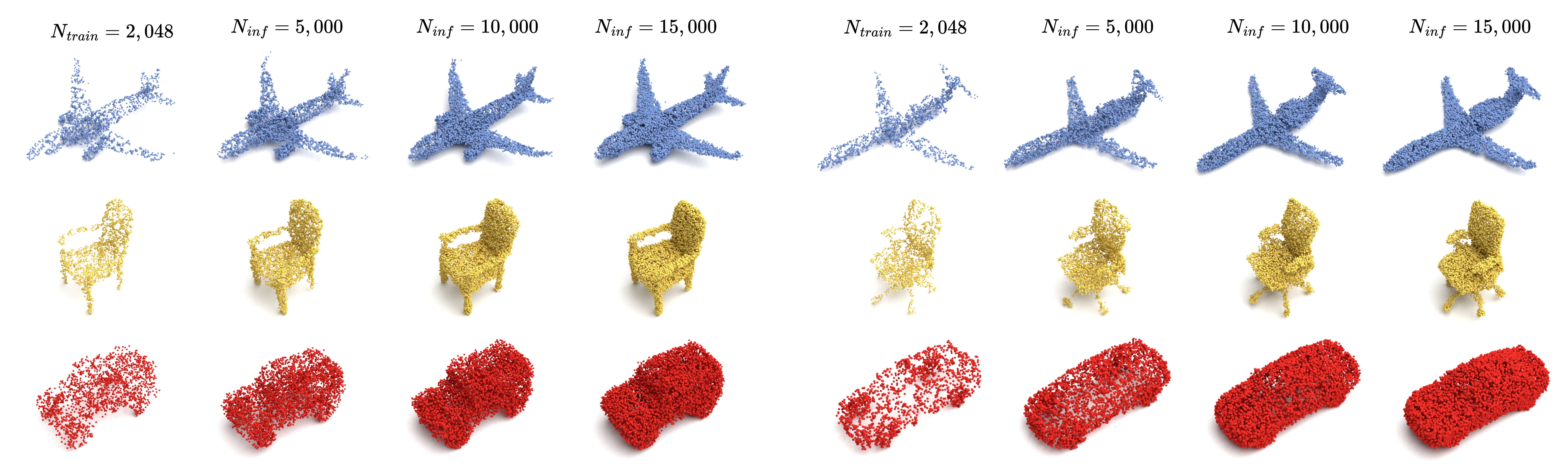}
    \caption{Generated samples using EMERGE and distribution alignment for varying resolutions.}
    \label{fig:multires_samples}
    \vspace*{-\baselineskip}
\end{figure}

\subsection{Super-Resolution Inference Pipeline}


While our distribution alignment provides robustness to density shifts, full denoising from $T$ to $0$ at target resolution $N_{inf}$ incurs severe computational overhead. We circumvent this with an efficient zero-shot inference pipeline inspired by iterative refinement \cite{sdedit}. Rather than generating high-resolution shapes from scratch, we reduce compute time by refining an upsampled low-resolution base shape:\looseness=-1

\begin{enumerate}[topsep = -2pt, leftmargin = 0.7cm]
    \item \textbf{Base Initialization:} The pipeline accepts an arbitrary base point cloud $\mathbf{X}_{base} \in \mathbb{R}^{N_{train} \times 3}$ (synthesized natively or provided as pre-generated).
    \item \textbf{Deterministic Upsampling:} The base point cloud is upsampled via point duplications to the target resolution $N_{inf}$, yielding an initial dense cloud $\mathbf{X}_{dense}\in\mathbb{R}^{N_{inf}\times 3}$.
    \item \textbf{Forward Noise Injection:} To break the spatial symmetries of the overlapping points, we partially corrupt $\mathbf{X}_{dense}$ by simulating the forward diffusion process up to an intermediate timestep $t_{res}$, naturally perturbing and separating the coordinates.
    \item \textbf{Zero-Shot Refinement:} Finally, we denoise the corrupted point cloud from $t_{res}$ to $t=0$ utilizing our distribution alignment technique. Specifically, at each denoising step, we apply Propositions 1 and 2 to align the voxel-count distributions and to preserve the spatial receptive field, respectively, yielding the final high-resolution point cloud $\mathbf{X}_{hr}\in\mathbb{R}^{N_{inf}\times 3}$.
\end{enumerate}
The complete inference pipeline is presented in Algorithm~\ref{alg:inference_pipeline} (Appendix \ref{app:multires_algorithm_section}). Qualitative results demonstrating its effectiveness are visualized in Figure \ref{fig:multires_samples}, where, instead of full DDPM sampling, we denoise from $t_{res}=150$ using 30 DDIM  steps. Furthermore, Figure \ref{fig:multires} shows that inference at higher resolutions without distribution alignment induces geometric distortion and structural collapse, whereas our pipeline preserves the global topology and fine-grained details of the generated point cloud.\looseness=-1


\section{Conclusion}
In this work, we presented EMERGE, a fully $SE(3)$-equivariant graph-based diffusion model that addresses the critical limitations of prior frameworks for 3D point cloud generation. By incorporating novel canonical voxel pooling and global invariant feature attention, our architecture successfully models complex, hierarchical geometries while strictly preserving continuous spatial symmetries. Furthermore, we theoretically prove that density-induced distribution shifts can be isolated and negated, empowering our framework to achieve zero-shot, high-fidelity 3D generation at arbitrary inference resolutions. Quantitative and qualitative evaluations confirm that EMERGE achieves State-of-the-Art generation fidelity and converges in significantly fewer training epochs than existing methods.\looseness=-1

\begin{ack}
Research supported by the NVIDIA Academic Grant Program using a single Brev instance with 8xA100 gpus.
\end{ack}

\newpage

\bibliographystyle{plainnat} 
\bibliography{references}








\appendix

\section{\texorpdfstring{$\boldsymbol{SE(3)}$}{SE(3)} Equivariance of Voxel Pooling}
\label{equivariant-voxel-pooling-appendix}

\noindent Let  $\mathbf{1}_N \in \mathbb{R}^{N\times 1}$ and  $\mathbf{1}_M \in \mathbb{R}^{M\times 1}$  be column vectors of ones, and let $\mathbf{X}\in \mathbb{R}^{N \times 3}$ be a point cloud consisting of $N$ points. Let $\mathbf{g} \in SE(3)$ be a transformation parametrized by a rotation matrix $\mathbf{R} \in SO(3)$ and a translation column vector $\mathbf{t} \in \mathbb{R}^{3\times 1}$, such that its action on $\mathbf{X}$ is defined by $\mathbf{X}' = \mathbf{XR} + \mathbf{1}_{N}\mathbf{t}^{T}$.

\noindent Let $f: \mathbb{R}^{N \times 3} \rightarrow \mathbb{R}^{M \times 3}$  denote our proposed Canonical Frame Voxel Pooling Operation. We assert that $f$ is exactly equivariant to the Special Euclidean group $SE(3)$, satisfying:
\begin{equation}
    f(\mathbf{XR} + \mathbf{1}_{N}\mathbf{t}) = f(\mathbf{X})\mathbf{R} + \mathbf{1}_{M}\mathbf{t}^{T}.
\end{equation}

\noindent \textit{Proof}

\textbf{Step 1. Centering}\\
Let $\boldsymbol{\mu} \in\mathbb{R}^{1\times 3}$ be the centroid of the original point cloud $\mathbf{X}$, calculated as $\boldsymbol{\mu} = \frac{1}{N}\mathbf{1}_{N}^{T}\mathbf{X}$.  The centered point cloud is defined as $\mathbf{X}_{c} = \mathbf{X} - \mathbf{1}_{N}\boldsymbol{\mu}$.

After applying the transformation $\mathbf{g}$ to $\mathbf{X}$, the new centroid $\boldsymbol{\mu}'$ is:
\begin{equation}
    \boldsymbol{\mu}' = \frac{1}{N}\mathbf{1}_{N}^{T}\left(\mathbf{X}\mathbf{R} + \mathbf{1}_{N}\mathbf{t}^{T} \right) = \left(\frac{1}{N}\mathbf{1}_{N}^{T}\mathbf{X}\right)\mathbf{R} + \frac{1}{N}\left( \mathbf{1}_{N}^{T}\mathbf{1}_{N}\right)\mathbf{t}^{T} = \boldsymbol{\mu}\mathbf{R} + \mathbf{t}^{T}.\label{transformed_centrod}
\end{equation}
The centered coordinates $\mathbf{X}_{c}^{'}$ for the transformed point cloud $\mathbf{X}'$, based on Eq.(\ref{transformed_centrod}) become:
\begin{equation}
    \mathbf{X}_{c}^{'} = \mathbf{X}' - \mathbf{1}_{N}\boldsymbol{\mu}' = (\mathbf{XR} + \mathbf{1}_{N}\mathbf{t}^{T}) - \mathbf{1}_{N}(\boldsymbol{\mu}\mathbf{R} + \mathbf{t}^{T}) = (\mathbf{X} - \mathbf{1}_{N}\boldsymbol{\mu})\mathbf{R} = \mathbf{X}_{c}\mathbf{R}.\label{transformed_centroid}
\end{equation}
From Eq.(\ref{transformed_centroid}) we observe that the translation term $\mathbf{t}$ is perfectly canceled, rendering $\mathbf{X}_{c}$ striclty \textbf{invariant} to translations and strictly \textbf{equivariant} to rotations. 

\noindent \textbf{Step 2. Covariance and Frame Extraction}\\ 
The covariance matrix $\mathbf{C}$ is computed as $\mathbf{C} = \mathbf{X}_{c}^{T}\mathbf{X}_{c}\in \mathbb{R}^{3\times 3}$. The eigendecomposition of $\mathbf{C}$ yields an orthogonal matrix of eigenvectors $\mathbf{V}\in\mathbb{R}^{3\times 3}$. Because any eigenvector $\mathbf{v}_{j}$ and its negation $-\mathbf{v}_{j}$ satisfy the eigenvalue equation equally, the basis vectors spanning the eigenspaces are mathematically unique only up to a sign. Consequently, evaluating the eigenvectors yields an arbitrary raw frame $\mathbf{V}\in O(3)$.

For the transformed centered points $\mathbf{X}_{c}^{'}$ the corresponding covariance matrix $\mathbf{C}'$, based on Eq.(\ref{transformed_centroid}) is equal to:
\begin{equation}
    \mathbf{C}' = \left(\mathbf{X}_{c}^{'}\right)^{T} \left(\mathbf{X}_{c}^{'}\right) = \left(\mathbf{X}_{c}\mathbf{R}\right)^{T} \left(\mathbf{X}_{c}\mathbf{R}\right) = \mathbf{R}^{T}\mathbf{X}_{c}^{T}\mathbf{X}_{c}\mathbf{R} = \mathbf{R}^{T}\mathbf{C}\mathbf{R}.
\end{equation}
From the above equation it follows that the principal axes of $\mathbf{C}'$ are exactly the principal axis of $\mathbf{C}$ rotated by $\mathbf{R}^{T}$. Let $\mathbf{V}'\in O(3)$ denote the frame extracted for the transformed system, which is equal to $\mathbf{R}^{T}\mathbf{V}$. Due to the inherent sign ambiguity of the eigenvectors, the relationship between the two extracted frames is given by:
\begin{equation}
    \mathbf{V}' = \mathbf{R}^{T}\mathbf{V}\mathbf{D},\label{transformed_v}
\end{equation}
where $\mathbf{D} = \text{diag}(d_{1}, d_{2}, d_{3})$ with $d_{j} \in \{-1, 1\}$ represents the arbitrary sign inversions along each principal axis.

To eliminate $\mathbf{D}$ in Eq.(\ref{transformed_v}) and striclty guarantee a unique, rotationally equivariant frame $\mathbf{V}^{\ast}\in SO(3)$, we resolve the two inherent ambiguities of the eigenspace.

\textbf{Remark on Degenerate Eigenvalues:} We acknowledge that in cases of perfect rotational symmetry (where eigenvalues are degenerate, e.g., $\lambda_1 \approx \lambda_2 \approx \lambda_3$), the eigenvectors become ill-defined and strict equivariance could theoretically be compromised. However, such instances are statistically insignificant for finite point clouds. Empirically, these degenerate covariance matrices are uniquely encountered when processing pure Gaussian noise at $t \approx T$, a state in which the arbitrary geometric alignment does not negatively influence the downstream generation.

\textbf{2A. Spatial Extent Disambiguation}\\
We first project the points onto the raw frame to obtain tentative coordinates $\mathbf{\hat{Y}} = \mathbf{X}_{c}\mathbf{V}$. Let $y_{max, j} = \max_{i}\left(\mathbf{\hat{Y}}_{i, j}\right)$ and $y_{min, j} = \min_{i}\left(\mathbf{\hat{Y}}_{i, j}\right)$. We define the correction sign matrix,
\begin{equation}
    \mathbf{S} = \text{diag}\left(s_{1}, s_{2}, s_{3} \right), \text{ where } s_{j} = \text{sgn}\left( y_{max, j} + y_{min, j}\right), \text{ with } \text{sgn}(0) = 1.
\end{equation}
This forces the axis to point toward the spatial maximum of the data. The oriented frame is then $\mathbf{\tilde{V}} = \mathbf{V}\mathbf{S}$.

For the transformed point cloud, the tentative coordinates are:
\begin{equation}
   \hat{\mathbf{Y}'} = \mathbf{X}_{c}'\mathbf{V}'=\left(\mathbf{X}_{c}\mathbf{R} \right)\left(\mathbf{R}^{T}\mathbf{VD} \right) = \mathbf{X}_{c}\left(\mathbf{RR}^{T}\right)\mathbf{VD} = \mathbf{X}_{c}\mathbf{VD} = \mathbf{\hat{Y}}\mathbf{D}.
\end{equation}
Because $\hat{\mathbf{Y}}' = \hat{\mathbf{Y}}\mathbf{D}$, the projected coordinates of the transformed point cloud along the $j$-th axis are given by $\hat{y}'_{i,j} = d_j \hat{y}_{i,j}$ for all points $i=1,\dots,N$. Since $d_j \in \{-1, 1\}$, this operation represents either an identity mapping or a strict reflection along the axis.If $d_j = 1$, the spatial extents remain identical: $y'_{max, j} = y_{max, j}$ and $y'_{min, j} = y_{min, j}$. Conversely, if $d_j = -1$, the axis is inverted, which geometrically causes the maximum and minimum values of the set to swap and negate: $y'_{max, j} = -y_{min, j}$ and $y'_{min, j} = -y_{max, j}$. In both scenarios, the sum of the spatial extremes factors out $d_j$ exactly:
\begin{equation}
    y'_{max, j} + y'_{min, j} = d_j \left(y_{max, j} + y_{min, j}\right).\end{equation}
Consequently, when computing the new correction heuristic $s'_{j} = \text{sgn}(y'_{max, j} + y'_{min, j})$, we obtain:
\begin{equation}
    s'_j = \text{sgn}\left( d_j \left(y_{max, j} + y_{min, j}\right) \right).
\end{equation}
Because $d_j$ is a non-zero scalar ($\pm 1$), it trivially factors out of the signum function, yielding $s'_j = d_j s_j$. In matrix notation, this establishes the strict identity $\mathbf{S}' = \mathbf{D}\mathbf{S}$. Applying this identity to the transformed oriented frame, we get:
\begin{equation}
    \tilde{\mathbf{V}}' = \mathbf{V}'\mathbf{S}' = \left(\mathbf{R}^{T}\mathbf{VD}\right)\left( \mathbf{DS}\right) = \mathbf{R}^{T}\mathbf{V}\mathbf{D}^{2}\mathbf{S}.
\end{equation}
Since $d_j \in \{-1, 1\}$, it is strictly true that $\mathbf{D}^2 = \mathbf{I}$. Thus:
\begin{equation}
    \tilde{\mathbf{V}}' = \mathbf{R}^T \mathbf{V} \mathbf{I} \mathbf{S} = \mathbf{R}^T (\mathbf{V} \mathbf{S}) = \mathbf{R}^T \tilde{\mathbf{V}}.\label{new_v}
\end{equation}
From Eq. (\ref{new_v}), we can clearly see that the arbitrary solver ambiguity $\mathbf{D}$ cancels out.

\textbf{2B. Reflections}\\
While $\tilde{\mathbf{V}}$ is deterministic, it is only guaranteed that $\tilde{\mathbf{V}} \in O(3)$. If $\det(\tilde{\mathbf{V}}) = -1$, the frame represents a left-handed coordinate system. Projecting into this frame would induce a reflection, destroying the geometric chirality of the input point cloud.

To strictly enforce $\mathbf{V}^* \in SO(3)$, we compute a chirality correction matrix $\mathbf{F} = \text{diag}(1, 1, \det(\tilde{\mathbf{V}}))$, which inverts the least significant principal axis if the frame is reflected. The final canonical frame is $\mathbf{V}^* = \tilde{\mathbf{V}} \mathbf{F}$. By definition, $\det(\mathbf{V}^*) = \det(\tilde{\mathbf{V}}) \cdot \det(\mathbf{F}) = \det(\tilde{\mathbf{V}})^2 = 1$.

We now prove this correction is invariant to $\mathbf{R}$. The determinant of the rotated oriented frame is:
\begin{equation}
    \det(\tilde{\mathbf{V}}') = \det(\mathbf{R}^T \tilde{\mathbf{V}}) = \det(\mathbf{R}^T)\det(\tilde{\mathbf{V}}).
\end{equation}
Because the applied rotation $\mathbf{R} \in SO(3)$, its determinant is exactly $1$. Thus, $\det(\tilde{\mathbf{V}}') = \det(\tilde{\mathbf{V}})$, meaning the exact same correction matrix is generated: $\mathbf{F}' = \mathbf{F}$. The final transformed frame resolves to:\looseness=-1
\begin{equation}
    (\mathbf{V}^*)' = \tilde{\mathbf{V}}' \mathbf{F}' = (\mathbf{R}^T \tilde{\mathbf{V}})\mathbf{F} = \mathbf{R}^T (\tilde{\mathbf{V}}\mathbf{F}) = \mathbf{R}^T \mathbf{V}^*.\label{new_v_ast}
\end{equation}
\textbf{Step 3. Canonical Projection}\\
The points are projected into the uniquely resolved canonical frame via $\mathbf{Y} = \mathbf{X}_c \mathbf{V}^*$. Substituting the transformed components $\mathbf{X}'_c$ and $(\mathbf{V}^*)'$ from Eqs. (\ref{transformed_centroid}) and (\ref{new_v_ast}) we get:
\begin{equation}
    \mathbf{Y}' = \mathbf{X}'_c (\mathbf{V}^*)' = (\mathbf{X}_c \mathbf{R})(\mathbf{R}^T \mathbf{V}^*) = \mathbf{X}_c (\mathbf{R} \mathbf{R}^T) \mathbf{V}^* = \mathbf{X}_c \mathbf{V}^* = \mathbf{Y}.
\end{equation}
This proves the canonical representation $\mathbf{Y}$ is strictly invariant to any $SE(3)$ transformation. Consequently, any deterministic voxel pooling function $\phi(\mathbf{Y})$ applied to this space yields an invariant set of pooled nodes $\mathbf{Y}_{\text{pool}} = \phi(\mathbf{Y})$, such that $\mathbf{Y}'_{\text{pool}} = \mathbf{Y}_{\text{pool}}$.

\textbf{Step 4. Unprojection}\\
To return the pooled nodes to world space, the canonical points are unprojected via $\mathbf{P} = \mathbf{Y}_{\text{pool}} (\mathbf{V}^*)^T + \mathbf{1}_{M}\boldsymbol{\mu}$.

For the transformed pipeline, the final output coordinates $\mathbf{P}'$ are:
\begin{equation}
    \mathbf{P}' = \mathbf{Y}'_{\text{pool}} ((\mathbf{V}^*)')^T + \mathbf{1}_{M}\boldsymbol{\mu}'.
\end{equation}

Substituting the derived identities $\mathbf{Y}'_{\text{pool}} = \mathbf{Y}_{\text{pool}}$, $(\mathbf{V}^*)' = \mathbf{R}^T \mathbf{V}^*$, and $\boldsymbol{\mu}' = \boldsymbol{\mu} \mathbf{R} + \mathbf{t}^{T}$:
\begin{align}
    \mathbf{P}' & = \mathbf{Y}_{\text{pool}} (\mathbf{R}^T \mathbf{V}^*)^T + \mathbf{1}_{M}(\boldsymbol{\mu} \mathbf{R} + \mathbf{t}^{T})\nonumber \\[5pt]
    \mathbf{P}' &= \mathbf{Y}_{\text{pool}} (\mathbf{V}^*)^T \mathbf{R} + \mathbf{1}_{M}\boldsymbol{\mu} \mathbf{R} + \mathbf{1}_{M}\mathbf{t}^{T}.
\end{align}
Factoring out the right-multiplied rotation matrix $\mathbf{R}$:
\begin{align}
    \mathbf{P}' &= (\mathbf{Y}_{\text{pool}} (\mathbf{V}^*)^T + \mathbf{1}_{M}\boldsymbol{\mu})\mathbf{R} + \mathbf{1}_{M}\mathbf{t}^{T} = \mathbf{P} \mathbf{R} + \mathbf{1}_{M}\mathbf{t}^{T}.
\end{align}
The output coordinates $\mathbf{P}'$ of the voxel pooling operation applied to the transformed input $\mathbf{X}\mathbf{R}+\mathbf{t}$ algebraically map to the exact transformation of the original output $\mathbf{P}\mathbf{R}+\mathbf{t}$. Therefore, the function $f$ rigorously satisfies $SE(3)$ equivariance.
 
\hfill $\blacksquare$

\section{Invariant Geometric Features and Proof of Equivariance}
\label{invariant-features-appendix}

\subsection{Definition of Invariant Features}
To capture the local topology around a point $\mathbf{x}_i$, we consider its $k$-nearest neighbors $\mathcal{N}(i)$. Let $\mathbf{r}_{ji} = \mathbf{x}_j - \mathbf{x}_i$ denote the relative distance vectors. The local covariance matrix $\mathbf{C}_i \in \mathbb{R}^{3 \times 3}$ is defined as:\looseness=-1
\begin{equation}
\mathbf{C}_i = \frac{1}{|\mathcal{N}(i)|} \sum_{j \in \mathcal{N}(i)} \mathbf{r}_{ji}^T \mathbf{r}_{ji}.    
\end{equation}
Through eigen-decomposition, we extract the eigenvalues $\lambda_1 \ge \lambda_2 \ge \lambda_3 \ge 0$ and their corresponding eigenvectors $\mathbf{V} = [\mathbf{v}_1, \mathbf{v}_2, \mathbf{v}_3]$. From these, we construct our 13-dimensional invariant feature vector $\mathbf{f}_{inv, i}$, which consists of two groups:

\paragraph{Eigenvalue-based features (7 dimensions):}
These describe the intrinsic shape of the local neighborhood.
\begin{enumerate}[topsep = 0pt, leftmargin = 0.7cm]
    \item \textbf{Sum of eigenvalues:} $\sum_{k=1}^{3} \lambda_k = \lambda_1 + \lambda_2 + \lambda_3$
    \item \textbf{Omnivariance:} $(\lambda_1 \lambda_2 \lambda_3)^{1/3}$
    \item \textbf{Eigenentropy:} $-\sum_{k=1}^3 \lambda_k \ln(\lambda_k + \epsilon)$
    \item \textbf{Linearity:} $(\lambda_1 - \lambda_2) / \lambda_1$
    \item \textbf{Planarity:} $(\lambda_2 - \lambda_3) / \lambda_1$
    \item \textbf{Sphericity:} $\lambda_3 / \lambda_1$
    \item \textbf{Change of curvature:} $\lambda_3 / \sum_{k=1}^{3} \lambda_{k}$
\end{enumerate}

\paragraph{Absolute spatial moments (6 dimensions):}
These capture the distribution of points projected onto the principal axes. For each principal axis $w \in \{1, 2, 3\}$, we compute the first and second-order absolute moments:
\begin{enumerate}
    \item[8-10.] \textbf{First-order moments:} $\frac{1}{|\mathcal{N}(i)|} \sum_{j \in \mathcal{N}(i)} \left| \mathbf{r}_{ji} \mathbf{v}_w \right|$
    \item [11-13.] \textbf{Second-order moments:} $\frac{1}{|\mathcal{N}(i)|} \sum_{j \in \mathcal{N}(i)} \left( \mathbf{r}_{ji} \mathbf{v}_w \right)^2$
\end{enumerate}

\subsection{Formal Proof of SE(3) Invariance}
We assert that the feature vector $\mathbf{f}_{inv, i}$ is strictly invariant to any SE(3) transformation applied to the input point cloud.

\textit{Proof}

\noindent Let  $\mathbf{1}_N \in \mathbb{R}^{N\times 1}$  be the column vector of ones, and $\mathbf{X}\in \mathbb{R}^{N \times 3}$ a point cloud consisting of $N$ points. Let $\mathbf{g} \in SE(3)$ be a transformation parametrized by a rotation matrix $\mathbf{R} \in SO(3)$ and a translation column vector $\mathbf{t} \in \mathbb{R}^{3\times 1}$, such that its action on $\mathbf{X}$ is defined by $\mathbf{X}' = \mathbf{XR} + \mathbf{1}_{N}\mathbf{t}^{T}$.

\textbf{1. Translation Invariance}

The relative vectors for the transformed points become:
\begin{equation}
\mathbf{r}'_{ji} = \mathbf{x}'_j - \mathbf{x}'_i = (\mathbf{x}_j \mathbf{R} + \mathbf{t}^{T}) - (\mathbf{x}_i \mathbf{R} + \mathbf{t}^{T}) = (\mathbf{x}_j - \mathbf{x}_i) \mathbf{R} = \mathbf{r}_{ji} \mathbf{R}.   
\end{equation}

The translation term $\mathbf{t}$ is perfectly canceled. Thus, all subsequent operations on $\mathbf{r}'_{ji}$ are inherently translation-invariant.

\textbf{2. Covariance Rotation}

The covariance matrix of the transformed neighborhood is:
\begin{align}
    \mathbf{C}'_i & = \frac{1}{|\mathcal{N}(i)|} \sum_{j \in \mathcal{N}(i)} (\mathbf{r}'_{ji})^T (\mathbf{r}'_{ji}) = \frac{1}{|\mathcal{N}(i)|} \sum_{j \in \mathcal{N}(i)} (\mathbf{r}_{ji} \mathbf{R})^T (\mathbf{r}_{ji} \mathbf{R})\nonumber\\[5pt] 
    & = \mathbf{R}^T \left( \frac{1}{|\mathcal{N}(i)|} \sum_{j \in \mathcal{N}(i)} \mathbf{r}_{ji}^T \mathbf{r}_{ji} \right) \mathbf{R} = \mathbf{R}^T \mathbf{C}_i \mathbf{R}.
\end{align}

\textbf{3. Invariance of Eigenvalues}

Let $\mathbf{C}_i \mathbf{V} = \mathbf{V} \boldsymbol{\Lambda}$ be the eigen-decomposition of the original covariance matrix. For the transformed matrix $\mathbf{C}'_i$:
\begin{equation}
    \mathbf{C}'_i (\mathbf{R}^T \mathbf{V}) = (\mathbf{R}^T \mathbf{C}_i \mathbf{R}) (\mathbf{R}^T \mathbf{V}) = \mathbf{R}^T \mathbf{C}_i (\mathbf{R} \mathbf{R}^T) \mathbf{V}.
\end{equation}
Because $\mathbf{R} \in SO(3)$, $\mathbf{R} \mathbf{R}^T = \mathbf{I}$. Therefore:
\begin{equation}
    \mathbf{C}'_i (\mathbf{R}^T \mathbf{V}) = \mathbf{R}^T (\mathbf{C}_i \mathbf{V}) = \mathbf{R}^T (\mathbf{V} \boldsymbol{\Lambda}) = (\mathbf{R}^T \mathbf{V}) \boldsymbol{\Lambda}.
\end{equation}
This demonstrates that the transformed covariance matrix $\mathbf{C}'_i$ has the exact same eigenvalues $\boldsymbol{\Lambda}' = \boldsymbol{\Lambda}$, while its eigenvectors rotate equivariantly as $\mathbf{V}' = \mathbf{R}^T \mathbf{V}$. Because the first 7 dimensions of $\mathbf{f}_{inv, i}$ are derived entirely from $\boldsymbol{\Lambda}$, they are strictly $SE(3)$-invariant.

\textbf{4. Invariance of Spatial Moments}

To compute the spatial moments, we project the transformed relative vectors $\mathbf{r}'_{ji}$ onto the transformed eigenvectors $\mathbf{V}'$:
\begin{equation}
    \mathbf{r}'_{ji} \mathbf{V}' = (\mathbf{r}_{ji} \mathbf{R}) (\mathbf{R}^T \mathbf{V}) = \mathbf{r}_{ji} (\mathbf{R} \mathbf{R}^T) \mathbf{V} = \mathbf{r}_{ji} \mathbf{V}.
\end{equation}
The rotation matrix perfectly cancels out ($\mathbf{R} \mathbf{R}^T = \mathbf{I}$), leaving the scalar projections exactly identical to those computed in the original frame. Consequently, the absolute moments (dimensions 8-13) are also strictly $SE(3)$-invariant. 

Since all 13 elements of $\mathbf{f}_{inv, i}$ are invariant, the vector itself is SE(3)-invariant. Therefore, multiplying the equivariant relative vectors by a function of $\mathbf{f}_{inv, i}$ in Equation (\ref{eq:egnn_f_inv}) preserves the overall SE(3)-equivariance of the coordinate update mechanism.

\section{Feature-wise Linear Modulation (FiLM)}
\label{appendix_film_layer}

Throughout our architecture, we frequently inject external conditioning information (such as the diffusion timestep $\mathbf{t}_{emb}$ or the voxel point counts) into the intermediate node features. To achieve this without disrupting the spatial equivariance of the coordinate updates, we utilize Feature-wise Linear Modulation (FiLM) layers \citep{film}.

Given an input feature matrix $\mathbf{H} \in \mathbb{R}^{N \times D}$ and a conditioning vector $\mathbf{c} \in \mathbb{R}^{D_{cond}}$ (which can be a global vector per shape or a specific value per node), the FiLM layer applies an affine transformation to the features. First, the conditioning vector is projected into a higher-dimensional space via a linear layer to generate scale ($\boldsymbol{\gamma}$) and shift ($\boldsymbol{\beta}$) parameters:
\begin{equation}
    [\boldsymbol{\gamma}, \boldsymbol{\beta}] = \mathbf{c}\mathbf{W}^T + \mathbf{b}.
\end{equation}
where $\mathbf{W} \in \mathbb{R}^{2D \times D_{cond}}$ is the learnable weight matrix, $\mathbf{b} \in \mathbb{R}^{2D}$ is the bias, and the resulting concatenated vector is split evenly along the feature dimension such that $\boldsymbol{\gamma}, \boldsymbol{\beta} \in \mathbb{R}^{D}$. 

The intermediate features are then modulated element-wise:
\begin{equation}
    \text{FiLM}(\mathbf{H}, \mathbf{c}) = \mathbf{H} \odot (1 + \boldsymbol{\gamma}) + \boldsymbol{\beta}.
\end{equation}
where $\odot$ denotes the Hadamard (element-wise) product. We formulate the scaling factor as $(1 + \boldsymbol{\gamma})$ rather than $\boldsymbol{\gamma}$ to act as a residual identity mapping when the network is initialized with small weights, ensuring stable training dynamics early in the diffusion process.

\section{Voxel Counts-Conditioning Normalization}

As introduced in Section \ref{sec:multiscale_egnn}, mapping continuous points to discrete spatial voxels naturally yields varying point densities per voxel. We utilize these point counts to condition the pooled features, providing the network with explicit structural density context at multiple resolutions. 

However, raw point counts $c_n$ for a pooled node $n$ can exhibit significant variance, with dense core regions aggregating orders of magnitude more points than sparse outliers. Directly feeding raw counts into the network can cause numerical instability and dominate the feature representations. To address this, we apply a two-step normalization process before the FiLM conditioning:
\begin{enumerate}[leftmargin=1.5em]
    \item \textit{Logarithmic Transformation:} We first compress the wide dynamic range of the counts using a smooth logarithmic transform to heavily penalize extreme outliers:
    \begin{equation}
        c'_n = \log(1 + c_n).
    \end{equation}
    \item \textit{Per-Shape Min-Max Normalization:} Point clouds within a dataset can vary drastically in their local density distributions. Normalizing the counts globally across a training batch would entangle the distinct topologies of different shapes. Therefore, we independently min-max normalize the log-counts strictly on a \textit{per-shape} basis. For a specific point cloud within the batch, we compute its minimum and maximum log-counts ($c'_{min, g}$ and $c'_{max, g}$) and normalize every pooled node $n$ as follows:
    \begin{equation}
        \tilde{c}_n = \frac{c'_n - c'_{min, g}}{c'_{max, g} - c'_{min, g} + \epsilon}.
    \end{equation}
    where $\epsilon = 10^{-8}$ is a small constant to prevent division by zero in geometrically uniform spaces. 
\end{enumerate}

This bounded, shape-normalized scalar $\tilde{c}_n \in [0, 1]$ is then passed as the 1-dimensional conditioning vector ($D_{cond}=1$) to the dedicated scale FiLM layer, which effectively modulates the $D$-dimensional pooled node features based on their relative structural density.

\section{Proofs for Zero-Shot Distribution Alignment}
\label{appendix:distribution-shift-proofs}
In this section, we provide the formal proofs for the distribution alignment technique introduced in Section \ref{sec:zero_shot_multiresolution}, which enable our model's zero-shot super-resolution generation capabilities.

\subsection{Proof of Scale-Count Expectation Alignment}
\label{appendix_scale_count_alignment}

\textbf{Proposition 1. (Scale-Count Alignment)}\\
 To preserve the expected input distribution of the count-conditioning FiLM layers, the raw inference voxel counts $c_n(N_{inf})$ must be linearly scaled by the inverse density ratio prior to normalization: $\hat{c}_n = c_n(N_{inf}) \cdot (N_{train} / N_{inf})$.
 
 \textit{Proof}
 
  Let the underlying 3D shape be modeled as a locally smooth 2D surface manifold $\mathcal{M}$ with total surface area $A$. This assumption naturally aligns with standard dataset construction practices, where point clouds are typically derived via uniform spatial sampling from the continuous surfaces of 3D meshes. We assume $N$ points are sampled uniformly over $\mathcal{M}$, yielding a global point density $\rho = N / A$.\looseness=-1
  
  During the $SE(3)$-equivariant multi-scale voxel pooling operation, the space is partitioned into canonical grids defined by the discrete voxel sizes in $\mathbf{V}_{vox} = \{v_0, v_1, \dots, v_M\}\subset \mathbb{R}_{>0}^{M+1}$. For any chosen scale $v_m \in \mathbf{V}_{vox}$, let $\mathcal{B}_v$ denote a specific voxel bin that intersects the manifold, enclosing a local surface area $A_{\mathcal{B}}$. Assuming a uniform local distribution, the expected number of points falling into this specific voxel $\mathcal{B}_v$ is given by the spatial integral of the density over $A_{\mathcal{B}}$:
  \begin{equation}
      \mathbb{E}[c_n(N)] = \int_{\mathcal{B}_v \cap \mathcal{M}} \rho \, dA = \rho A_{\mathcal{B}} = \frac{N}{A} A_{\mathcal{B}}.
  \end{equation}
  Let $N_{train}$ denote the number of points used during training, and $N_{inf}$ denote the target resolution at inference, where $N_{inf} > N_{train}$. The expected voxel counts for each state are:
  \begin{align}
      \mathbb{E}[c_n(N_{train})] &= \frac{N_{train}}{A} A_{\mathcal{B}},\label{training_expectation}\\
   \mathbb{E}[c_n(N_{inf})] &= \frac{N_{inf}}{A} A_{\mathcal{B}}.\label{inference_expectation}   
  \end{align}
  Substituting $A_{\mathcal{B}} / A$ from Eq. (\ref{training_expectation}) into Eq. (\ref{inference_expectation}) we get:
  \begin{equation}
  \mathbb{E}[c_n(N_{inf})] = \frac{N_{inf}}{N_{train}} \mathbb{E}[c_n(N_{train})].    
  \end{equation}
  To align the distributions before the logarithmic normalization and FiLM conditioning layers, we define the scaled inference count $\hat{c}$ as:
  \begin{equation}
    \hat{c}_n = c_n(N_{inf}) \cdot \frac{N_{train}}{N_{inf}}.   
  \end{equation}
  Taking the expectation of $\hat{c}_n$:
  \begin{align}
      \mathbb{E}[\hat{c}_n] &= \mathbb{E} \left[ c_n(N_{inf}) \cdot \frac{N_{train}}{N_{inf}} \right] = \frac{N_{train}}{N_{inf}} \mathbb{E}[c_n(N_{inf})] \nonumber\\[8pt]
      \mathbb{E}[\hat{c}_n] &= \frac{N_{train}}{N_{inf}} \left( \frac{N_{inf}}{N_{train}} \mathbb{E}[c_n(N_{train})] \right) = \mathbb{E}[c_n(N_{train})].
  \end{align}
  Thus, across all scales $v_{m} \in \mathbf{V}_{vox}$, the linear scaling aligns the expected input distribution of the FiLM layer exactly with the distribution observed during training, neutralizing the density shift.

  \hfill $\blacksquare$

  \subsection{Proof of Dilated $k$-NN Receptive Field Preservation}
\label{appendix_dilated_knn}
\textbf{Proposition 2 (Dilated $k$-NN Receptive Field Matching)}\\ Let $d = \lfloor N_{inf} / N_{train} \rfloor$ be the spatial dilation factor. A $k$-NN graph querying $k_{ext} = k \cdot d$ neighbors and subsampling every $d$-th neighbor perfectly preserves both the spatial receptive field radius $r_{train}$ and the message-passing cardinality $k$.

\textit{Proof}

 Let $r_k$ denote the physical distance to the $k$-th nearest neighbor for a point $\mathbf{x}_i, i = 1,\ldots, N$ on the surface manifold $\mathcal{M}$. Assuming the manifold is locally smooth, the local neighborhood forms a geodesic disk of area $\approx \pi r_k^2$. For the disk to enclose exactly $k$ points under a uniform density $\rho = N/A$, the expected area must satisfy:
 \begin{equation}
     \rho (\pi r_k^2) \approx k \implies \left(\frac{N}{A}\right) \pi r_k^2 \approx k.
 \end{equation}
 Solving for the physical radius $r_k$:
 \begin{equation}
     r_k(N) \approx \sqrt{\frac{k A}{\pi N}}.
 \end{equation}
 During training ($N = N_{train}$), the network optimizes its weights based on the structural context enclosed within the radius $r_{train}$:
 \begin{equation}
     r_{train} \approx \sqrt{\frac{k A}{\pi N_{train}}}.
 \end{equation}
 During inference ($N = N_{inf} = d \cdot N_{train}$), maintaining a standard $k$-NN graph causes the radius to shrink inversely proportional to the square root of the density ratio:
 \begin{equation}
     r_{inf, standard} \approx \sqrt{\frac{k A}{\pi (d \cdot N_{train})}} = \frac{r_{train}}{\sqrt{d}}.
 \end{equation}
 To restore the receptive field, the Dilated $k$-NN algorithm expands the initial query to $k_{ext} = k \cdot d$ neighbors. The physical radius encompassing these $k_{ext}$ points is:
 \begin{equation}
     r_{inf, dilated} \approx \sqrt{\frac{k_{ext} A}{\pi N_{inf}}} = \sqrt{\frac{(k \cdot d) A}{\pi (N_{train} \cdot d)}} = \sqrt{\frac{k A}{\pi N_{train}}} = r_{train}.
 \end{equation}
 Thus, the spatial boundary of the neighborhood is perfectly preserved.
 
 To complete the graph construction, the algorithm sorts the $k \cdot d$ points by distance and subsamples every $d$-th point. The number of selected neighbors $|\mathcal{N}(i)_{dilated}|$ is exactly:
 \begin{equation}
     |\mathcal{N}(i)_{dilated}| = \frac{k \cdot d}{d} = k
 \end{equation}
 Because the message-passing aggregations in the EGNN sum exactly $k$ terms, the magnitude and variance of the aggregated feature vectors remain perfectly in-distribution. 

 \hfill $\blacksquare$

It is also important to highlight that we apply this Dilated $k$-NN only to the EGNN that transforms the base point cloud, and we keep the standard $k$-NN for all the EGNNs that operate on voxel-pooled versions. This is due to the fact that voxel sizes (and therefore also their locations and density) are predefined as model hyperparameters, so they never change even when $N_{inf} \neq N_{train}$, so there is no need for Dilated $k$-NN on these layers.

\section{Super-Resolution Inference Pipeline Pseudo-code}
\label{app:multires_algorithm_section}
In this section, we provide the complete pseudo-code for the zero-shot super-resolution inference pipeline introduced in Section~\ref{sec:zero_shot_multiresolution}. Algorithm~\ref{alg:inference_pipeline} details the step-by-step procedure, including the base initialization, deterministic upsampling, forward noise injection, and the application of our parameter-free distribution alignment during the reverse denoising process.

\begin{algorithm}[t!]
\caption{Zero-Shot Super-Resolution Inference Pipeline}
\label{alg:inference_pipeline}
\begin{algorithmic}[1]
\renewcommand{\baselinestretch}{1.3}\selectfont
\Require Base point cloud $\mathbf{X}_{base} \in \mathbb{R}^{N_{train} \times 3}$, Target resolution $N_{inf}$, Intermediate timestep $t_{res}$, Pretrained model $f_\theta$, Noise schedule $\bar{\alpha}_{t}$
\Ensure High-resolution point cloud $\mathbf{X}_{hr} \in \mathbb{R}^{N_{inf} \times 3}$

\State \textbf{\textit{// Step 1: Base Initialization}}
\State Load or natively synthesize base point cloud $\mathbf{X}_{base}$

\State \textbf{\textit{// Step 2: Deterministic Upsampling}}
\State $d \gets \lfloor N_{inf} / N_{train} \rfloor$ \Comment{Calculate spatial dilation factor}
\State $\mathbf{X}_{dense} \gets \text{DuplicatePoints}(\mathbf{X}_{base}, d)$ \Comment{Yields shape of size $N_{inf} \times 3$}

\State \textbf{\textit{// Step 3: Forward Noise Injection}}
\State $\boldsymbol{\epsilon} \sim \mathcal{N}(\mathbf{0}, \mathbf{I})$
\State $\mathbf{X}_{t_{res}} \gets \sqrt{\bar{\alpha}_{t_{res}}} \mathbf{X}_{dense} + \sqrt{1 - \bar{\alpha}_{t_{res}}} \boldsymbol{\epsilon}$ \Comment{Diffuse to intermediate timestep}

\State \textbf{\textit{// Step 4: Zero-Shot Refinement with Distribution Alignment}}
\State $\mathbf{X}_{t} \gets \mathbf{X}_{t_{res}}$
\For{$t = t_{res}, t_{res}-1, \dots, 1$} \Comment{Can utilize DDPM or DDIM sampling}
    
    \State \textbf{\textit{Distribution Alignment Step:}}
    \State For all voxel scales $v_m$, scale raw counts: $\hat{c}_n \gets c_n(N_{inf}) \cdot (N_{train} / N_{inf})$
    \State Construct Dilated $k$-NN graph (dilation factor $d$) only on base point cloud scale
    
    \State \textbf{\textit{Model Prediction:}}
    \State $\tilde{\mathbf{X}}_0 \gets f_\theta(\mathbf{X}_{t}, t)$ \Comment{Denoise using aligned graph \& scaled counts}
    
    \State \textbf{\textit{Sampling Update:}}
    \State Compute $\mathbf{X}_{t-1}$ from $\mathbf{X}_{t}$ and $\tilde{\mathbf{X}}_0$ using standard diffusion transitions
\EndFor

\State $\mathbf{X}_{hr} \gets \mathbf{X}_{0}$
\State \Return $\mathbf{X}_{hr}$
\end{algorithmic}
\vspace{\baselineskip}
\end{algorithm}

\section{Higher-Resolution Distribution-Shift of the Core Architecture}
\label{appendix_architecture_immunity}

In Section \ref{sec:zero_shot_multiresolution}, we demonstrated that aligning the voxel-count distribution and preserving the $k$-NN receptive field eliminates the two primary bottlenecks for zero-shot super-resolution. Here, we also show that the remaining operations in the EMERGE architecture are natively invariant to the total point count $N$, ensuring no hidden distribution shifts occur when scaling to $N_{inf} > N_{train}$.

\paragraph{EGNN Message Passing Cardinality}
The EGNN updates for coordinates and features rely on aggregating messages over a local neighborhood $\mathcal{N}(i)$. As proven in Appendix \ref{appendix_dilated_knn}, our Dilated $k$-NN algorithm strictly guarantees that the inference neighborhood size remains exactly $|\mathcal{N}(i)_{dilated}| = k$. Consequently, both the coordinate update, which computes an average over neighbors:
\begin{equation}
    \Delta\mathbf{x}_i^l = \frac{1}{k} \sum_{j \in \mathcal{N}(i)_{dilated}} (\mathbf{x}_i^l - \mathbf{x}_j^l) \phi_x\left(\tilde{\mathbf{m}}_{ij}, \mathbf{f}_{inv, i}\right),
\end{equation}
and the feature update, which computes a sum over neighbors:
\begin{equation}
    \Delta\mathbf{h}_i^l = \phi_h\left(\mathbf{h}_i^l, \sum_{j \in \mathcal{N}(i)_{dilated}} \tilde{\mathbf{m}}_{ij}\right),
\end{equation}
aggregate exactly $k$ terms. Because the cardinality and spatial bounds of the neighborhood match the training phase precisely, the variance and expected magnitude of these aggregations remain completely unshifted.

\paragraph{Scale-Agnostic Voxel Pooling}
During the Multi-Scale Point Cloud Processing phase, the node features are obtained by mean-pooling the features of all points that fall within a given voxel. Let a given voxel $n$ enclose $c_n$ points. The pooled feature representation at voxel scale $v_m$ is $\frac{1}{c_n}\sum_{j=1}^{c_n} \mathbf{h}_{j, m}$. As the total points $N$ increases to $N_{inf}$, the number of points within the voxel increases proportionally by $\rho = N_{inf}/N_{train}$. However, because mean-pooling fundamentally computes the expected feature value within that local spatial volume, evaluating it with a denser point cloud simply acts as a higher-resolution Monte Carlo estimate of the continuous spatial expectation. The expected magnitude of the pooled feature is mathematically independent of $c_n$.

\paragraph{Global Inducing Point Attention}
The Global Invariant Feature Attention module utilizes a fixed set of $C$ inducing points, where $C \ll N_{train}$. During the compression stage, these inducing points act as queries to aggregate information from the $N$ node features via Cross-Attention:
\begin{equation}
    \text{Attention}(\mathbf{Q}, \mathbf{K}, \mathbf{V}) = \text{softmax}\left(\frac{\mathbf{Q}\mathbf{K}^T}{\sqrt{D}}\right)\mathbf{V}.
\end{equation}
The $\text{softmax}$ function normalizes the attention weights across the variable sequence length $N$, ensuring the weights inherently sum to $1$ regardless of the number of keys. As $N$ increases, the individual attention weights become proportionally smaller, but their sum remains constant. Therefore, the output of the attention mechanism is always a convex combination of the value vectors $\mathbf{V}$. Much like the local mean-pooling operation above, increasing $N$ simply provides a denser, finer-grained sampling of the underlying continuous shape manifold, leaving the expected magnitude and variance of the global context vectors perfectly aligned with the training distribution. We also don't utilize any positional embeddings for the point features $\mathbf{H}$ (which would obviously incur distribution shifts at different resolution).

\section{Extended Experimental Setup and Implementation Details}
\label{app:implementation_details}

In this section, we provide the complete architectural and hyperparameter details necessary to reproduce our experiments, expanding upon the summary provided in Section \ref{sec:experiments}.

\paragraph{Architecture} Our architecture utilizes $N_{\textit{B}} = 6$ main blocks, each interleaving a local Multi-Scale EGNN with a Global Inducing Point Attention module. Within the Multi-Scale EGNN, each of the $7$ resolution branches (the base resolution plus $M=6$ voxel scales) is independently processed by a Multi-Layer EGNN consisting of $N_E = 2$ sequential message-passing layers ($k=5$). The Inducing Point Attention module comprises 128 learnable inducing points and 8 attention heads. Further details are presented in Table \ref{tab:hyperparameters}. 

\begin{table}[!h]
    \centering
    \def\arraystretch{1.2}
    \caption{Complete hyperparameters and architectural configurations for EMERGE.}
    \label{tab:hyperparameters}
    \begin{tabular}{llc}
        \toprule
        \textbf{Category} & \textbf{Hyperparameter} & \textbf{Value} \\
        \midrule
        \midrule
        \multirow{5}{*}{\textbf{Architecture}} 
        & Hidden Dimension ($D$) & 512 \\
        & Multi-Scale EGNN Blocks ($N_{B}$) & 6 \\
        & EGNN Layers per Block ($N_{E}$) & 2 \\
        & Global Attention Heads & 8 \\
        & Inducing Points & 128 \\
        & \# Params & $285$M\\
        \midrule
        \multirow{3}{*}{\textbf{Graph \& Pooling}} 
        & $k$-Nearest Neighbors ($k$) & 5 \\
        & Training Resolution ($N_{train}$) & 2,048 \\
        & Voxel Sizes $\{v_{1}, \ldots v_{M} \}$  & \{0.2, 0.4, 0.65, 1.0, 2.8, 4.0\} \\
        \midrule
        \multirow{3}{*}{\textbf{Diffusion}} 
        & Timesteps ($T$) & 1000 \\
        & Noise Schedule ($\beta$) & Linear ($10^{-4}$ to $0.02$) \\
        & Min-SNR Clipping ($\gamma$) & 5 \\
        \midrule
        \multirow{8}{*}{\textbf{Training}}
        & Optimizer & AdamW \\ 
        & Batch Size & 160 \\
        & Learning Rate & $1e-4$ \\
        & Learning Rate  Scheduler & Cosine \\ 
        & Warmup Steps & 300 \\
        \cmidrule{2-3} 
        & Epochs (\textit{Airplane}) & $3,100$ \\
        & Epochs (\textit{Chair}) & $2,600$ \\
        & Epochs (\textit{Car}) & $7,000$ \\
        & Epochs (\textit{55-Class}) & $2,100$\\
        \bottomrule
        \bottomrule
    \end{tabular}
\end{table}

A fundamental characteristic of our $SE(3)$-equivariant architecture is that it generates point clouds in arbitrary spatial orientations, rather than defaulting to the canonical poses of the training set. Because standard distance metrics such as CD and EMD are strictly pose-dependent, directly evaluating unaligned point clouds would artificially inflate the error margins. To ensure a fair and mathematically rigorous comparison against non-equivariant baselines, we apply Principal Component Analysis (PCA) to both the generated and ground-truth shapes prior to evaluation. By aligning the principal axes of the point clouds, we isolate the true geometric fidelity and structural quality of the generated manifolds, independent of their spatial orientation.

All experiments were conducted on a 4xH100 GPU node, with a total VRAM of 256GB, 512GB RAM, and 2x Intel Sapphire Rapids 8460Y+ (80 cores in total). In terms of computational efficiency, a full 100-step DDIM generation takes $\sim41$ seconds, whereas our zero-shot super-resolution to 15,000 points given a base shape (with 30 inference steps) requires just $\sim34$ seconds.

\paragraph{Local and Global Multi-Scale Pooling} For the local geometric interactions within the EGNN layers, we construct a $k$-nearest neighbor graph utilizing $k=5$. To establish a robust hierarchical geometry representation, our equivariant voxel pooling procedure is applied across 6 distinct spatial resolutions. Specifically, we utilize the following voxel pooling sizes: $\{u_{1}, \ldots, u_{M}\} = \{0.2, 0.4, 0.65, 1.0, 2.8, 4.0\}$, where $M = 7$.

\section{Min-SNR Weighting Strategy Formulation}
\label{app:min_snr}

As established in our methodology, our model is designed to predict the clean data $\mathbf{X}_0$ rather than the added noise $\boldsymbol{\epsilon}$. By reparameterizing the standard simplified diffusion objective in Eq. (\ref{diffusion_noise_loss}), the base loss function for $\mathbf{X}_0$-prediction takes the equivalent form:
\begin{equation}
    \mathcal{L}_{\mathbf{X}_{0}} = \mathbb{E}_{t\sim\mathcal{U}[1,T], \mathbf{X}_{0}\sim q(\mathbf{X}_{0}), \boldsymbol{\epsilon}\sim\mathcal{N}(\mathbf{0},\mathbf{I})} \left[ \frac{\bar{\alpha}_{t}}{1-\bar{\alpha}_{t}} ||\mathbf{X}_{0} - f_{\theta}(\mathbf{X}_{t},t)||_{2}^{2} \right].
\end{equation}

However, directly optimizing this objective can lead to severe training instabilities and slow convergence rates due to conflicting optimization directions across different diffusion timesteps. To mitigate this, we utilize the Min-SNR-$\gamma$ weighting strategy \citep{Hang2023EfficientDT}. By applying this truncation strategy, the final training objective optimized by our network becomes:
\begin{equation}
    \mathcal{L}_{\mathbf{X}_0, \textit{Min-SNR}} = \mathbb{E}_{t \sim \mathcal{U}[1,T], \mathbf{X}_0 \sim q(\mathbf{X}_0), \boldsymbol{\epsilon} \sim \mathcal{N}(\mathbf{0},\mathbf{I})} \left[ \min(\operatorname{SNR}(t), \gamma) ||\mathbf{X}_0 - f_\theta(\mathbf{X}_t, t)||_2^2 \right],\label{x0_loss_function}
\end{equation}
where $\text{SNR}(t) = \frac{\bar{\alpha}_t}{1-\bar{\alpha}_t}$ and we set $\gamma=5$ \citep{Hang2023EfficientDT}.

\begin{figure}[t]
    \centering
    \includegraphics[width=\textwidth]{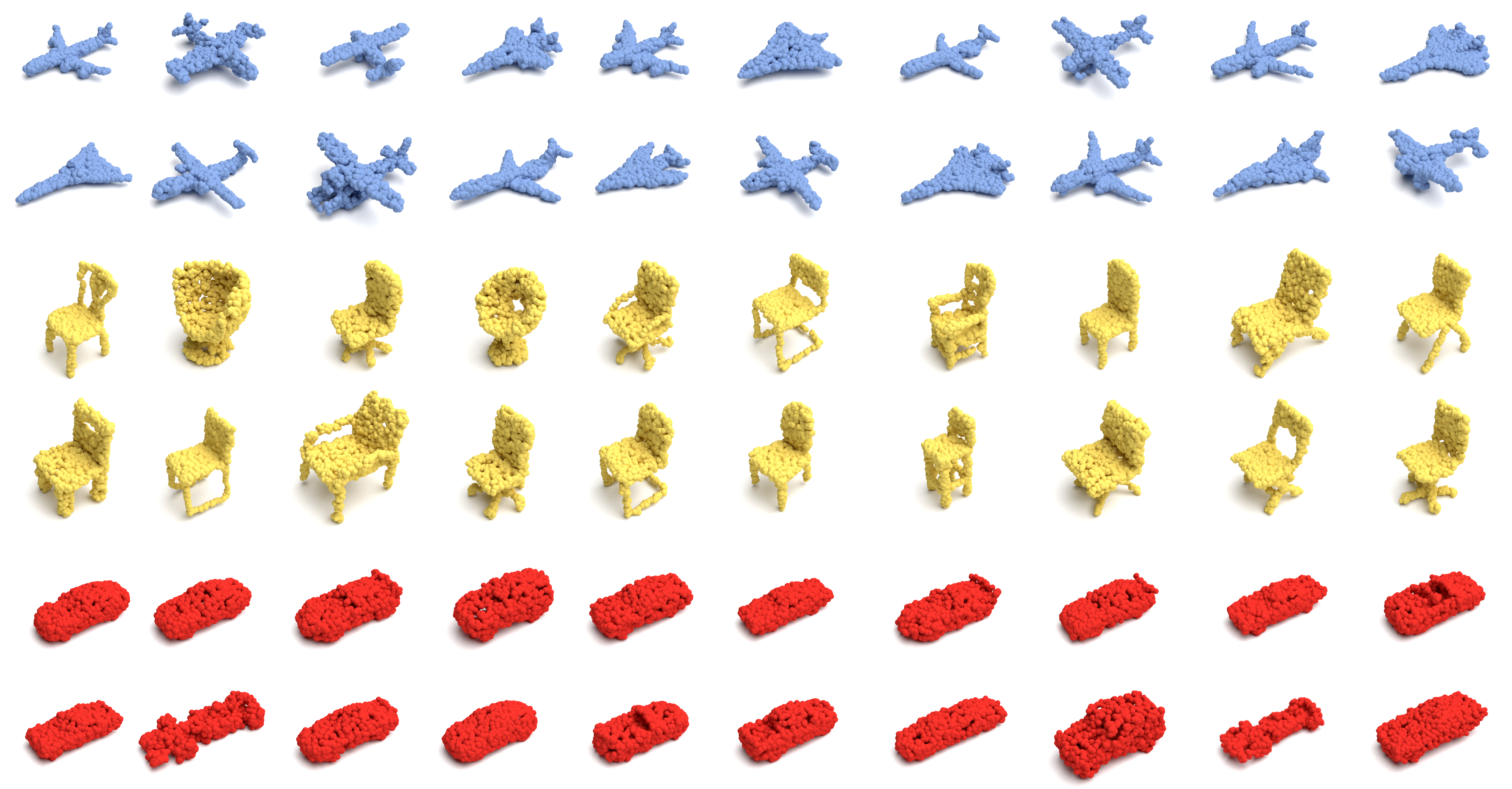}
    \caption{Generated samples using EMERGE for the Shapenet classes: \textit{Airplane}, \textit{Car}, and \textit{Chair}.}
    \label{fig:appendix_extra_quals}
\end{figure}

\begin{figure}[t]
    \centering
    \includegraphics[width=0.95\textwidth]{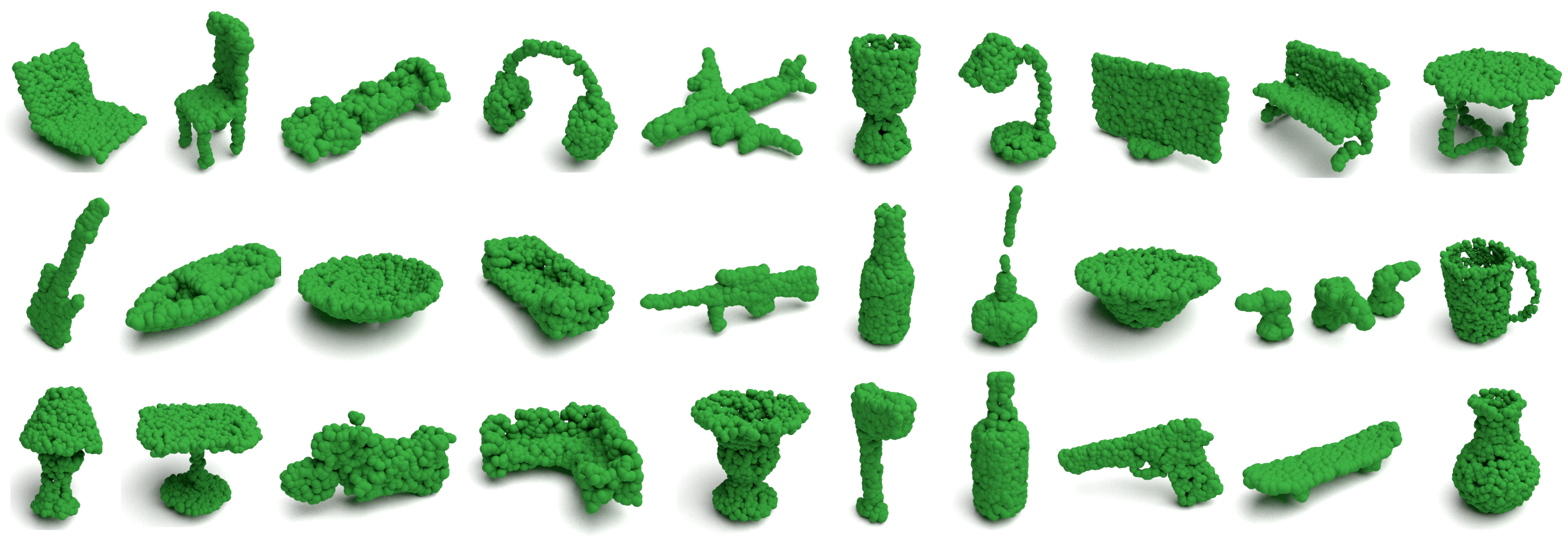}
    \caption{Samples generated by EMERGE for all 55 categories of ShapeNet.}
    \label{fig:55_class_appendix}
\end{figure}

\begin{figure}[!t]
    \centering
    \includegraphics[width=\textwidth]{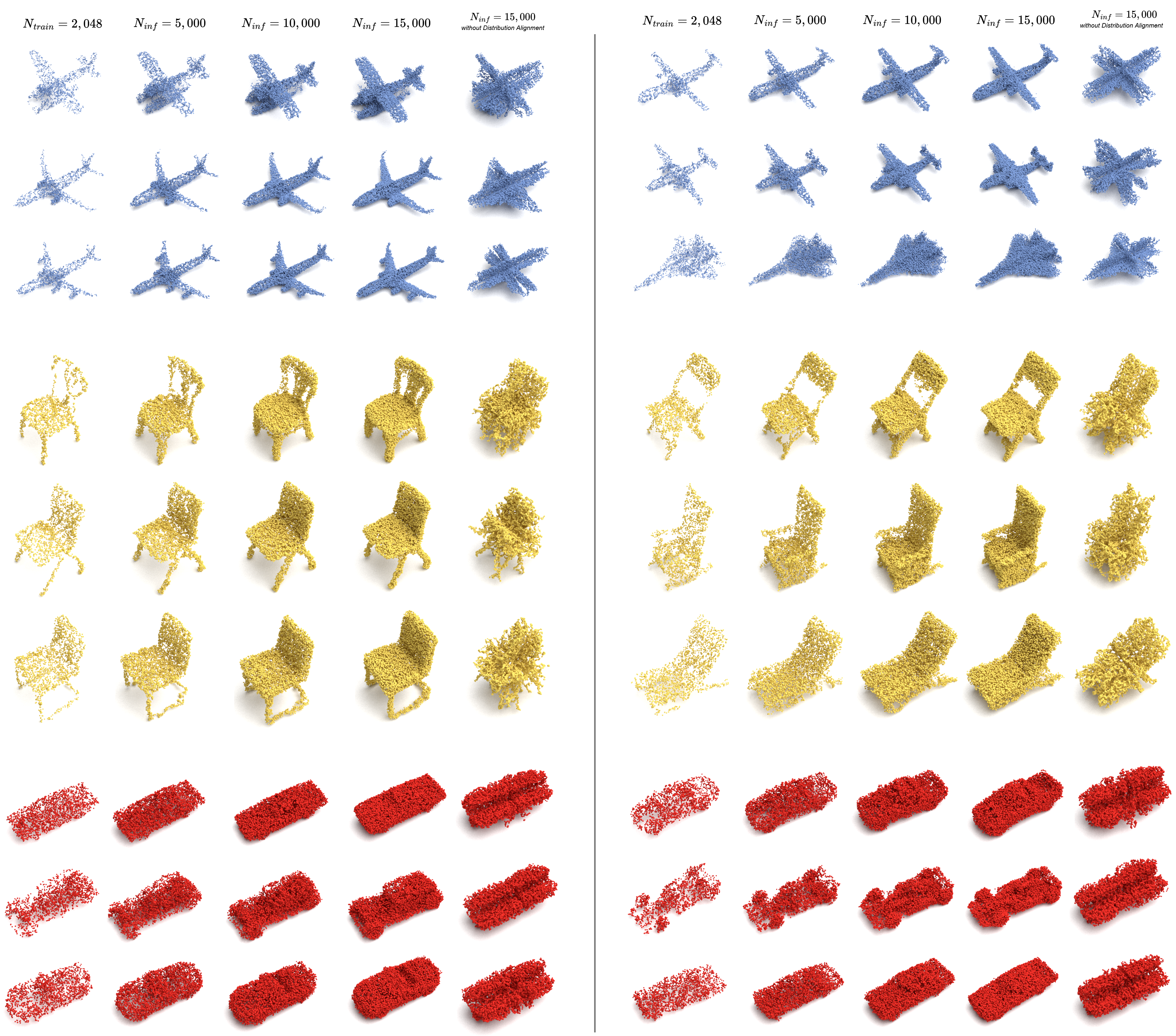}
    \caption{Generated samples using EMERGE, with and without distribution alignment for varying resolutions, for the Shapenet classes: \textit{Airplane}, \textit{Car}, and \textit{Chair}. }
    \label{fig:appendix_extra_multires_quals}
\end{figure}

\section{Extended Qualitative Results}
\label{appendix:extended_results}

In this section, we provide additional qualitative results to further demonstrate the generative capacity, diversity, and scalability of EMERGE.

\subsection{Standard Benchmark Categories}
In Figure \ref{fig:appendix_extra_quals}, we present extended generation samples for the three core ShapeNet benchmark categories (\textit{Airplane}, \textit{Car}, and \textit{Chair}). These results highlight the ability of our $SE(3)$-equivariant backbone to consistently produce high-fidelity geometries with smooth surfaces and sharp structural details.

\subsection{55-Class Generation}
 Figure \ref{fig:55_class_appendix} illustrates extended generation samples for all 55 classes of ShapeNet. Upon scrutinizing this figure, we deduce that EMERGE successfully synthesizes a wide variety of objects, ranging from dense, volumetric shapes (e.g., sofas, vessels) to intricate, thin structures (e.g. lamps, firearms, guitars, earphones etc.).

\subsection{Zero-Shot Super-Resolution}
\begin{figure}[!b]
    \centering
    \includegraphics[width=\textwidth]{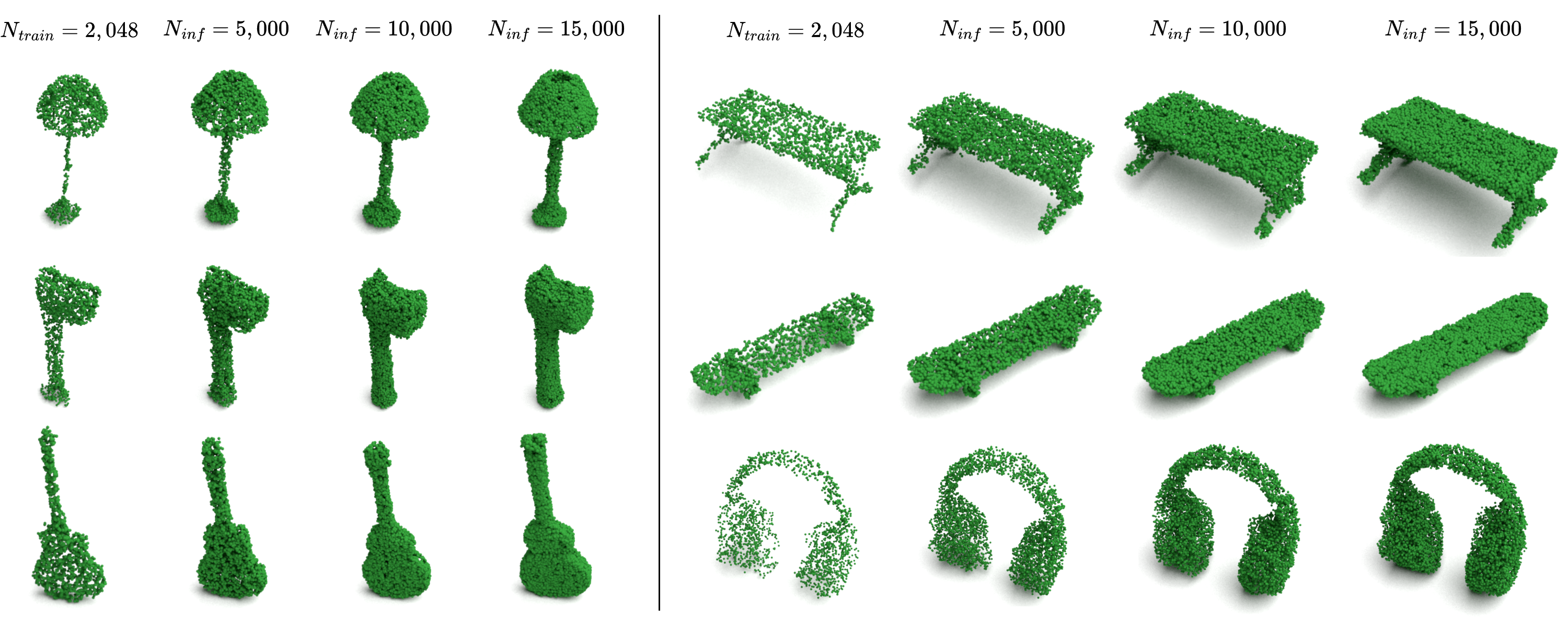}
    \caption{Generated samples using EMERGE and distribution alignment for varying resolutions, for all 55 classes of ShapeNet.}
    \label{fig:app_super_res_55}
\end{figure}

Finally, to further validate our proposed parameter-free distribution alignment technique, Figures \ref{fig:appendix_extra_multires_quals} and \ref{fig:app_super_res_55} provide additional examples of EMERGE performing zero-shot super-resolution for the three categories \textit{Airplane}, \textit{Car}, and \textit{Chair}, and for all 55 categories of ShapeNet, respectively. In Figure \ref{fig:appendix_extra_multires_quals}, we also showcase several super-resolution attempts using EMERGE without our proposed distribution alignment technique, which results in complete structural collapse of the point cloud, further proving that analytically resolving these receptive field shifts is essential for scaling inference beyond the training resolution.

\begin{figure}[t]
    \centering
    \includegraphics[height=0.98\textheight]{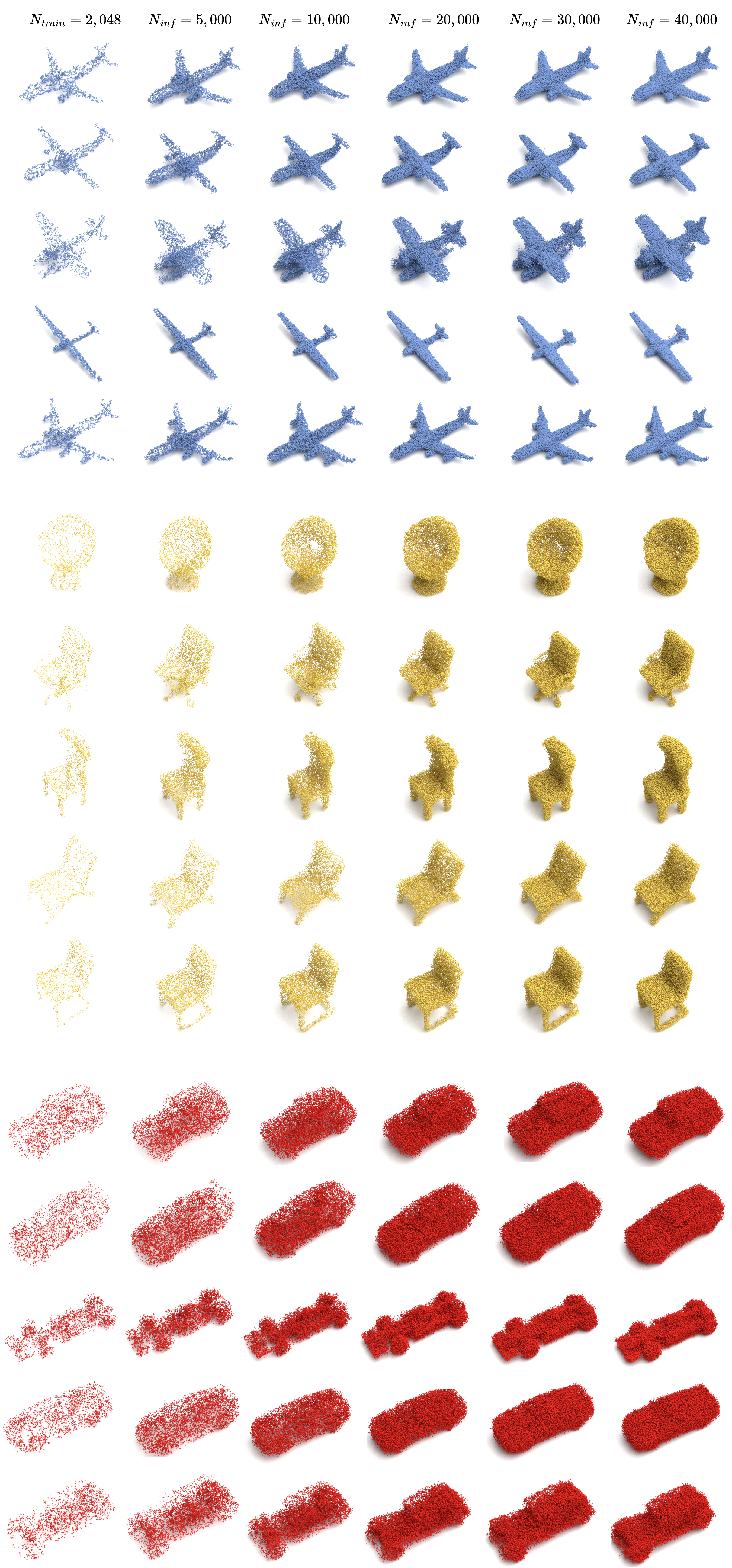}
    \caption{Extended zero-shot super-resolution results for resolutions up to 40,000 points ($N_{inf} = 40,000$).}    
    \label{fig:multires_40k}
\end{figure}

To further demonstrate the scalability of our distribution alignment technique, Figure~\ref{fig:multires_40k} showcases zero-shot super-resolution scaled up to $N_{inf} = 40,000$ points. This proves that EMERGE maintains strict geometric stability even at exceptionally high density ratios.

\section{Limitations}
\label{appendix:limitations}

While EMERGE demonstrates strong sample efficiency and high-fidelity generation, its current implementation incurs certain computational bottlenecks. Unlike widely adopted architectures that benefit from years of highly specialized, hardware-level GPU optimization, our framework relies on complex geometric operations that lack equivalent native support. Specifically, the dynamic construction of the $k$-Nearest Neighbor graph at each layer and the analytical eigenvector computations required for both our canonical frame extraction and invariant feature representations introduce noticeable latency. Consequently, although EMERGE requires vastly fewer total epochs to converge, the wall-clock time per individual training step and inference pass is comparatively higher than that of standard diffusion baselines.

\clearpage
\newpage

\end{document}